\pdfoutput=1
\documentclass{article}
\PassOptionsToPackage{sort,numbers,compress}{natbib}

\usepackage[final,eandd]{neurips_2026}

\usepackage[utf8]{inputenc}
\usepackage[T1]{fontenc}

\usepackage{hyperref}
\hypersetup{pdftitle={MMTA: Benchmarking Multimodal Temporal Analysis with Time Series, Text, and Vision}, pdfauthor={Ziyang Zhang, Shenyi Li, Yilin Wang, Ziyun Cui, Bowen Zhou, Wen Wu, Chao Zhang}}
\usepackage{url}
\usepackage{booktabs}
\usepackage{array}
\usepackage{amsfonts}
\usepackage{amsmath}
\usepackage{amssymb}
\usepackage{nicefrac}
\usepackage{microtype}
\usepackage{xcolor}
\usepackage{graphicx}
\usepackage{multirow}
\usepackage{tcolorbox}
\usepackage{listings}
\tcbuselibrary{breakable, skins}

\lstdefinestyle{jsonstyle}{
  basicstyle=\ttfamily\footnotesize,
  breaklines=true,
  columns=fullflexible,
  frame=single,
  rulecolor=\color{black!20},
  framerule=0.4pt,
  backgroundcolor=\color{black!2},
  showstringspaces=false
}

\tcbset{
  myexample/.style={
    colback=black!2,
    colframe=black!35,
    boxrule=0.6pt,
    arc=2pt,
    left=6pt,right=6pt,top=6pt,bottom=6pt
  }
}

\title{MMTA: Benchmarking Multimodal Temporal Analysis with Time Series, Text, and Vision}

\author{Ziyang Zhang\textsuperscript{1,2}\thanks{Equal contribution.} \quad
  Shenyi Li\textsuperscript{2}\footnotemark[1] \quad
  Yilin Wang\textsuperscript{2} \quad
  Ziyun Cui\textsuperscript{1} \\
  {\bfseries Bowen Zhou\textsuperscript{2} \quad
  Wen Wu\textsuperscript{2} \quad
  Chao Zhang\textsuperscript{1,2}\thanks{Corresponding author.}} \\[4pt]
  \textsuperscript{1}Tsinghua University \quad
  \textsuperscript{2}Shanghai Artificial Intelligence Laboratory \\[2pt]
  \texttt{ziyang-z24@mails.tsinghua.edu.cn}, \texttt{cz277@tsinghua.edu.cn} \\
  \texttt{\{lishenyi, wuwen\}@pjlab.org.cn}
}

\begin{document}

\maketitle

\begin{abstract}
Time-series analysis is increasingly studied in multimodal settings, yet most benchmarks pair temporal signals with text or with visualizations derived from the same signal. This leaves open whether multimodal large language models (MLLMs) can integrate time series with real-world visual evidence. We introduce \textbf{MMTA} (\textbf{m}ulti\textbf{m}odal \textbf{t}emporal \textbf{a}nalysis), a benchmark of tri-modal tasks with diverse cross-modal complementarity patterns among time series, text, and vision. MMTA spans nine domains -- robotics, autonomous driving, human action, manufacturing, AR assistance, healthcare, recommendation, finance, and climate -- under \emph{video$+$TS} and \emph{image$+$TS} layouts, with a 3{,}524-sample test split covering classification and forecasting. The benchmark standardizes heterogeneous real-world sources into a shared schema with domain-specific prompts, provenance documentation, and success-weighted metrics that penalize both wrong answers and failed generations. Evaluating eight open-source and proprietary MLLMs reveals a shared bottleneck: representing time series as text creates very long token sequences and weakens fine-grained alignment with visual evidence, especially in long-video settings. We then introduce \textbf{TimeOmni-v}, a multimodal LLM extension with a dedicated time-series encoder, temporal positional alignment, modality-aware token layout, and a forecasting head. TimeOmni-v improves long-video classification and time-series inference efficiency, showing that MMTA not only exposes representation and grounding failures but also motivates targeted model design.\footnote{Project page: \url{https://windowso.github.io/MMTA/}. Data: \url{https://huggingface.co/datasets/OpenTSLab/MMTA-v1.0}. Code for TimeOmni-v and baseline evaluation: \url{https://github.com/OpenTSLab/TimeOmni-v}.}

\end{abstract}

\section{Introduction}
\label{sec:intro}

\begin{figure}[t]
    \centering
    \includegraphics[width=1\linewidth]{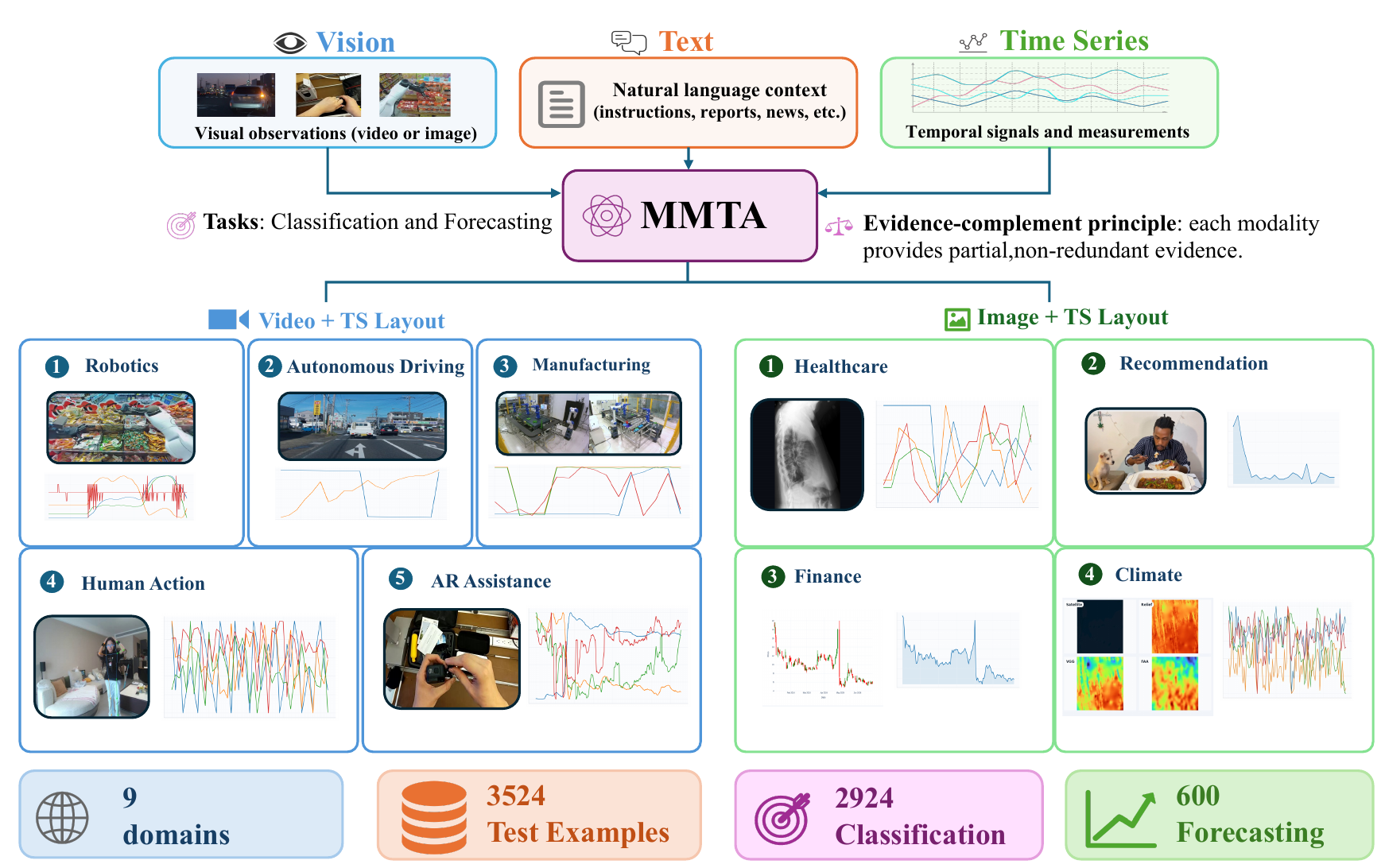}
    \caption{Overview of MMTA. The benchmark unifies nine real-world domains -- Robotics, Autonomous Driving, Human Action, Manufacturing, AR Assistance, Healthcare, Recommendation, Finance, and Climate -- under two cross-modal layouts (video $+$ time series and image $+$ time series). Vision provides real-world context, text provides semantic context, and time series provides quantitative dynamics; which modality dominates varies across domains.}
    \label{fig:dataset_fig}
\end{figure}

Time-series analysis, in both forecasting and classification, has long underpinned decision making in domains such as economics, meteorology, healthcare, and industrial monitoring. Recent work has therefore moved beyond unimodal temporal signals toward formulations that pair time series with natural language~\citep{wang2025itformer, chen2025mtbench, williamscontext, liutime, kong2025time, Zhou2025MoTime, wu2025scits}. This direction is motivated by a simple limitation: numerically similar temporal patterns can mean different things depending on the task, domain, and event context. Text helps by exposing instructions, reports, metadata, or external events that are not recoverable from the sequence alone.

Text, however, is only one part of the multimodal temporal setting. Many real-world decisions also depend on visual evidence that is not a rendering of the time series. Vision encodes physical and semantic context such as objects, scenes, actions, medical imagery, product appearance, or geophysical structure; time series provides high-resolution quantitative dynamics; and text supplies semantic context, events, reports, or instructions. These sources are structurally different and often only jointly informative. For example, as illustrated in Figure~\ref{fig:dataset_fig} for the robotics domain (case taken from the AgiBotWorld dataset~\cite{AgiBotWorld2026}), deciding whether a robot completed an episode correctly draws on the language instruction to establish the goal, the egocentric video to follow the operation, and the joint and end-effector traces to verify fine-grained execution. Existing benchmarks do not systematically test whether current MLLMs can perform this kind of evidence integration over visual, textual, and temporal streams.

We address this gap with \textbf{MMTA} (\textbf{M}ulti\textbf{m}odal \textbf{T}emporal \textbf{A}nalysis), a real-world benchmark that places vision, text, and time series side by side. MMTA spans nine domains: robotics, autonomous driving, human action, manufacturing, AR assistance, healthcare, recommendation, finance, and climate. Each domain is exported under one of two cross-modal layouts, video plus time series, or image plus time series. The test split contains 3{,}524 samples and covers both classification and forecasting. MMTA is a collection of tri-modal tasks with diverse cross-modal complementarity patterns: depending on the domain, a modality may be decisive, supporting, or nearly redundant. MMTA is intended as an evaluation substrate for studying temporal grounding, modality reliance, efficient time-series representation, and robust reasoning under missing or weakly aligned evidence, rather than only as a leaderboard.

Our systematic evaluation of eight representative open-source and proprietary MLLMs on MMTA exposes a common failure mode: serializing time series as text produces excessively long token sequences and offers no mechanism to align temporal samples with co-occurring visual evidence. We address this bottleneck with \textbf{TimeOmni-v}, a multimodal LLM extension designed for long time series in multimodal temporal tasks. TimeOmni-v attaches a dedicated time-series encoder to an existing multimodal LLM, gives time-series tokens a temporal positional structure, and arranges them relative to the visual stream. This design improves long-video classification and inference efficiency, while the remaining difficulty of multimodal forecasting highlights open problems surfaced by MMTA.

Our contributions are summarized as follows.
\begin{itemize}
    \item We introduce \textbf{MMTA}, a real-world collection of tri-modal tasks with diverse cross-modal complementarity patterns among vision, text, and time series, across nine domains and two layouts, video$+$TS and image$+$TS.
    \item We standardize heterogeneous sources into a shared evaluation suite with a 3{,}524-sample test split, classification and forecasting tasks, domain-specific prompts, provenance and license documentation, human and mechanical quality audits, and success-weighted metrics that penalize both incorrect answers and unparsable outputs.
    \item We characterize the difficulty of MMTA through eight representative MLLMs, identifying a common failure mode in which serializing time series as text creates excessive token length and weak visual-temporal alignment.
    \item We introduce \textbf{TimeOmni-v}, a multimodal LLM extension with a dedicated time-series encoder, temporal positional alignment, modality-aware token layout, and forecasting head. TimeOmni-v improves long-video classification and inference efficiency, can be fine-tuned on long-TS domains where the same backbone with textualized time series runs out of memory, and also runs on InternVL3.5-8B on the four datasets tested.
\end{itemize}

\section{Related Work}

\paragraph{Multimodal time-series datasets and benchmarks.}
Multimodal time-series datasets have grown along three lines. The first line pairs time series with free-form text such as news, reports, or task descriptions, as in MTBench~\citep{chen2025mtbench}, Context is Key~\citep{williamscontext}, TimeCAP~\citep{Lee2025TimeCAP}, Time-MMD~\citep{liutime}, NewsForecast~\citep{Wang2024NewsForecast}, Time-MQA~\citep{kong2025time}, and TimesX~\citep{Liu2026TimesX}; Time-IMM~\citep{chang2025timeimm} pairs irregular multivariate time series with text and adds images only as an extension in its appendix. The second line attaches rendered plots of the underlying signal~\citep{Yang2025TimeRA, Liu2024TimerBed, HORAI2025, MMTS2025}, so the image is a view of the series itself rather than an observation of the scene. The most recent line introduces real images alongside text and time series, exemplified by MoTime~\citep{Zhou2025MoTime}; even there, images are typically reduced to captions before consumption, only forecasting is supported, and domain coverage is concentrated on recommendation. MMTA differs from all three by using raw scene video or images as sample-level evidence, by aligning them with text and time series across nine real-world domains, and by covering both classification and forecasting.

\paragraph{Time-series-aware language models.}
Recent efforts to make language models natively consume time series stop short of fusing with vision, leaving an architectural slot that TimeOmni-v occupies. One line attaches a dedicated time-series encoder to a text LLM through token-level patch concatenation~\citep{chatts, timellm, quinlan2025chat, jia2024gpt4mts}; the resulting models reason over time series and text only, with no co-occurring visual stream. A separate line builds omni-modal LLMs with native text, image, audio, and video encoders~\citep{Qwen25Omni, Qwen3Omni}, yet none of them ships a time-series encoder, so long signals must still be serialized as digit strings inside the textual prompt. TimeOmni-v integrates time series into an omni-modal LLM as a first-class modality, on the same footing as vision and audio.

\section{The MMTA Benchmark}
\label{sec:benchmark}

MMTA is a collection of vision$+$TS$+$text tri-modal tasks with diverse cross-modal complementarity patterns, built from the openly available datasets in which the three modalities are natively present. Nine real-world domains are aligned into a single corpus under two cross-modal layouts (Figure~\ref{fig:dataset_fig}), \emph{video$+$TS} and \emph{image$+$TS}, and each domain is evaluated separately on a 3{,}524-sample test split (2{,}924 classification and 600 forecasting samples). The domains differ by an order of magnitude or more in time-series length (up to $5{\times}10^{3}$ steps), channel count (1 to 36), video duration (up to $155.7$\,s), and image count (1 to 7 per sample). Text is counted as a sample-level modality when it varies across samples and carries scene, event, or background information; seven domains carry such text, whereas the prompts of CoVLA and CUHK-X contain only the task description and the label space (Appendix~\ref{app:sample-text}). Provenance, licenses, and prompt templates are documented in Appendix~\ref{app:realworld_datasets}, and the release includes a Croissant metadata file with the Responsible AI fields.

\subsection{Video$+$TS Domains}
\label{sec:bench-video}

\paragraph{Robotics (AgiBotWorld).} The source is AgiBotWorld2026~\citep{AgiBotWorld2026}. Each instance is one Task Frame within a manipulation episode. The visual stream is the top-head egocentric clip covering the full task-frame interval, and the temporal stream is a 10-channel state trace combining six arm-position channels with four end-effector velocity channels. The text modality exposes the task instruction and the ordered step sequence. The model is asked to classify the dominant anomaly cause among four labels: \emph{paused}, \emph{meaningless operation}, \emph{fall}, or \emph{other}.

\paragraph{Autonomous Driving (CoVLA).} The source is the CoVLA Dataset~\citep{CoVLA2024}, which contains driving videos recorded on Japanese public roads with synchronized vehicle-state and lead-vehicle annotations. Each instance is a braking segment, defined as a continuous interval where the brake pedal stays pressed for at least six frames. The visual stream is the cropped dashcam clip over that interval, and the temporal stream is five synchronized channels: ego speed, ego acceleration, steering angle, lead-vehicle speed, and lead-vehicle acceleration. The model classifies the brake reason into three categories: lead-vehicle deceleration, ego turning, or other. The labels are derived by rules from upstream signals. The brake-pedal and turn-signal variables that locate the braking interval and define the ego-turning class are not among the five released channels, whereas lead-vehicle acceleration also enters the lead-vehicle-deceleration rule, so the label and the input time series overlap partially.

\paragraph{Human Action (CUHK-X).} The source is the CUHK-X HAR subset~\citep{CUHKX2024}. Each instance is one action trial. The visual stream is a synthetic skeleton-rendered clip, so pose evolution is encoded directly in the video rather than through raw RGB; the temporal stream is five wearable IMU streams, each containing three accelerometer channels and three gyroscope channels. The model classifies the action over 40 categories.

\paragraph{Manufacturing (Future Factories).} The source is the Future Factories Platform~\citep{Harik2024FutureFactories}. Each instance is one production cycle on an industry-grade assembly line. The visual stream is the on-site surveillance video captured by two synchronized cameras monitoring the assembly station; the temporal stream is 36 PLC channels (12 gripper signals plus 24 robot joint angles); and the text modality is a cycle report describing the process. Defects are introduced by manually removing components from a modular rocket assembly. The model classifies the cycle into one of four labels: \emph{Correct}, \emph{NoNose}, \emph{NoNose,NoBody2}, or \emph{NoNose,NoBody2,NoBody1}.

\paragraph{AR Assistance (HoloAssist).} The source is HoloAssist~\citep{HoloAssist2023}. Each instance is one coarse-grained action segment with a duration between 20 and 80 seconds. The visual stream is the egocentric RGB clip cropped to the segment, the temporal stream is nine synchronized HoloLens channels (three-axis gaze direction, three-axis head translation, three-axis gyroscope), and the text modality lists the ordered fine-grained actions inside the segment. The model is asked a binary correctness question: whether all contained fine-grained actions were executed correctly.

\subsection{Image$+$TS Domains}
\label{sec:bench-image}

\paragraph{Healthcare (MIMIC).} The sources are MIMIC-IV~\citep{Johnson2023MIMIC,PhysioNet-mimiciv-3.1}, MIMIC-CXR~\citep{johnson2019mimic,PhysioNet-mimic-cxr-jpg-2.1.0}, and MIMIC-IV-Note~\citep{PhysioNet-mimic-iv-note-2.2}. Each instance is a clinically anchored cohort point: either a chest X-ray acquisition or a fixed 48-hour mark after admission. The visual stream is the chest radiograph at the anchor; the temporal stream is six routinely recorded vital signs (heart rate, respiratory rate, SpO$_2$, temperature, NIBP systolic, NIBP diastolic) over the observation window; the text modality is the radiology report aligned to the same study, which records the radiologist's interpretation, clinical history, and prior-exam comparisons. Two prognosis-style tasks are evaluated: \emph{in-hospital mortality prediction} (predict whether the patient ultimately dies during the hospital stay using the first 48 hours of data) and \emph{discharge readiness prediction} (predict whether the patient should be discharged within the next 48 hours from a 48-hour lookback window).

\paragraph{Recommendation (PixelRec).} The source is the PixelRec dataset~\citep{Cheng2023PixelRec}, which captures video-consumption behaviors in lifestyle and entertainment categories. Each instance is one item. The visual stream is the item cover thumbnail; the temporal stream is the daily interaction count series (\textit{e.g.}, daily views or clicks); the text modality concatenates title, tag, and description. The model is asked to forecast the next-multi-day interaction volume, conditioning on the historical daily interactions, the cover image, and the metadata text.

\paragraph{Finance (FinMultiTime).} The source is the FinMultiTime dataset~\citep{Xu2025FinMultiTime}, instantiated on the S\&P 500 universe. Each instance is one asset paired with a half-year history window. The visual stream is the candlestick chart of the same window, which encodes the open, high, low, and close prices; the temporal stream is the daily closing-price series, so the two share only the close price; the text modality is a small set of recent news snippets selected strictly from before the prediction horizon. The model forecasts closing prices over the next 12 trading days, combining numeric trajectories with semantic drivers (events, sentiment) and visual technical patterns.

\paragraph{Climate (Terra).} The source is the Terra dataset~\citep{Chen2024Terra}, defined on a global 1$^{\circ}$ latitude--longitude grid. Each instance is one grid cell. The visual stream is a multi-image stack of remote-sensing and geophysical raster products (4 to 7 images per sample, including satellite imagery, geoid, free-air anomaly, magnetic anomaly, land--sea mask, relief, and vertical gravity gradient); the temporal stream is seven climate variables over the historical window (precipitation, temperature, surface pressure, relative humidity, specific humidity, wind speed, downward longwave radiation); the text modality is a geographic description of the cell that the Terra authors generated with an LLM from the per-cell metadata (coordinates, city, vegetation, climate, elevation). The model forecasts the next 20 daily values for all variables.

\subsection{Data Quality}
\label{sec:bench-quality}

Two components of MMTA are generated rather than annotated by humans: the CoVLA brake-reason labels, derived by rules in our pipeline, and the Terra descriptions, which the Terra authors generated with an LLM. AgiBotWorld, Future Factories, and HoloAssist keep the upstream label semantics through deterministic aggregation or label-space normalization. For each of the two generated components, 100 samples were checked independently by two annotators, and disagreements were resolved by adjudication. The CoVLA labels reach 96\% accuracy (Cohen's $\kappa=0.910$) and the Terra descriptions 93\% ($\kappa=0.687$). All 25{,}057 video$+$TS samples passed a mechanical audit of sample identity, time boundaries, and time-series coverage: 4 HoloAssist samples (0.016\%) do not fully cover the target window, and 1.42\% of the released frames lie outside it because clips are cut at keyframes. Image$+$TS domains are joined on deterministic keys. The audit protocol, error types, and per-domain results are given in Appendix~\ref{app:data-quality}.

\section{TimeOmni-v for Multimodal Temporal Analysis}
\label{sec:TimeOmni-v}

\begin{figure}[t]
    \centering
    \includegraphics[width=1\linewidth]{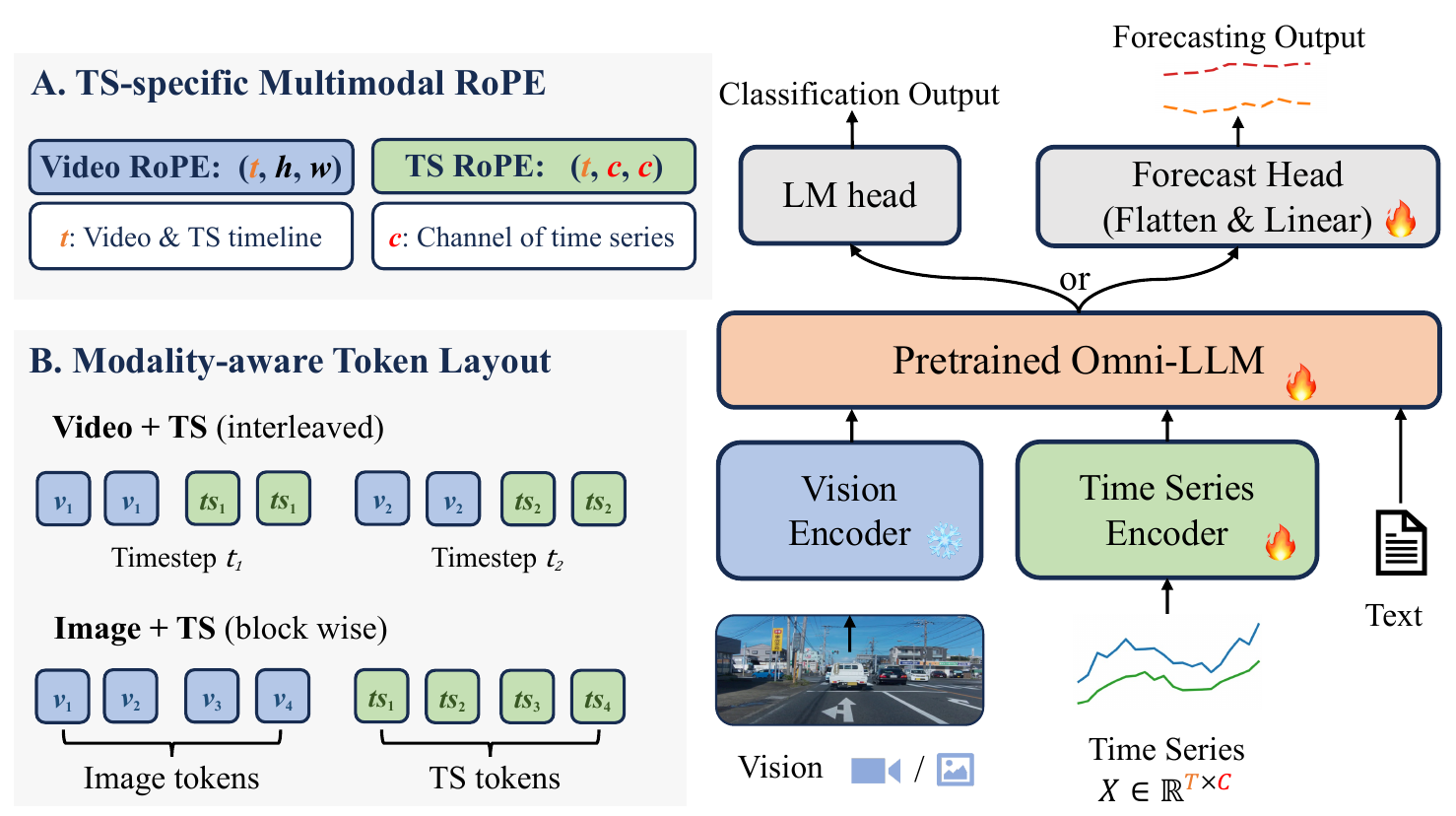}
    \caption{Overview of TimeOmni-v. Time series enters the multimodal LLM through a dedicated Chronos-2 encoder branch parallel to the vision encoder. TS tokens carry a TS-specific multimodal RoPE and are arranged relative to the visual stream via a modality-aware layout (interleaved with video, blocked against still images). A forecast head is attached for numerical prediction; the standard LM head is used for classification.}
    \label{fig:TimeOmni-v}
\end{figure}

To improve the efficiency of a multimodal LLM on long time series and to tighten its fusion with vision and text, we develop \textbf{TimeOmni-v} as an illustrative framework built on Qwen2.5-Omni~\citep{Qwen25Omni}. As illustrated in Figure~\ref{fig:TimeOmni-v}, TimeOmni-v comprises four components. A Chronos-2~\citep{Chronos2} encoder branch compresses the raw time-series stream into a compact token sequence. A TS-specific multimodal RoPE gives time series a 2D positional code aligned with the video timeline. A modality-aware token layout interleaves time-series tokens with video frames or blocks them next to still images. A lightweight forecast head emits multi-step numerical predictions in a single forward pass.

\subsection{Time-series modality integration}
\label{sec:TimeOmni-v-input}

The time-series modality enters the LLM through a Chronos-2~\citep{Chronos2} encoder branch parallel to the vision encoder. Chronos-2 is a pretrained time-series foundation model, trained on large-scale corpora via a forecasting objective. Given a multivariate input $\mathbf{X} \in \mathbb{R}^{T \times C}$ with $T$ time steps over $C$ channels, the encoder applies channel-independent patch tokenization along the time axis with patch length $P = 16$. It produces a patch embedding sequence $\mathbf{H} \in \mathbb{R}^{N \times D_{\text{enc}}}$ with $N = C \cdot \lceil T/P \rceil$ output tokens, where $D_{\text{enc}}$ is the Chronos-2 hidden dimension. A lightweight adapter then maps each patch embedding from $D_{\text{enc}}$ to the LLM hidden dimension $D_{\text{llm}}$, producing the TS token sequence $\mathbf{X}_{\text{ts}} \in \mathbb{R}^{N \times D_{\text{llm}}}$. $\mathbf{X}_{\text{ts}}$ is finally scattered into the LLM input embeddings.

To align time series with the visual stream both positionally and sequentially, TS tokens carry a TS-specific multimodal RoPE and follow a modality-aware layout. The RoPE uses layout $(t, c, c)$, with the temporal axis on the first channel and the channel index duplicated into the remaining two, replacing the $(t, h, w)$ axes used for video. When TS coexists with video, $t$ is rescaled onto the video timeline. A TS patch and the co-occurring video frame then share the same RoPE temporal value, which yields a zero relative-time offset under attention. The same TS tokens are time-interleaved with video frames when both are present, or grouped as a contiguous block next to still images. In the interleaved case, the video stream is split into per-frame token spans, and after each frame the $C$ TS tokens of the patch covering that frame's time interval are spliced in, so each frame is immediately followed by the cross-channel readings of the same instant.

\subsection{Forecast head for numerical prediction}
\label{sec:TimeOmni-v-head}

To produce multi-step numerical predictions in a single forward pass, TimeOmni-v attaches a lightweight forecast head on top of the LLM in place of autoregressive digit decoding. The head takes the last-layer hidden states $\mathbf{Z} \in \mathbb{R}^{W \times D_{\text{llm}}}$ over a fixed window of $W$ tokens anchored at the first TS token of the sample. A linear projection then maps $\mathbf{Z}$ to $\hat{\mathbf{Y}} \in \mathbb{R}^{L \times C}$ in normalized space, where $L$ is the prediction length and $C$ is the channel count. Per-sample input statistics are used to denormalize $\hat{\mathbf{Y}}$ to the original scale. The forecast head is attached only on forecasting tasks; classification reuses the standard LM head, so the same model serves both formats.

\section{Experimental Setup}

\subsection{Evaluated Models}
\label{sec:evaluated-models}
Both open-source and proprietary MLLMs are evaluated. The open-source set covers two families. The vision-language family contains InternVL3.5-4B~\citep{wang2025internvl3}, InternVL3.5-8B~\citep{wang2025internvl3}, Qwen3-VL-8B-Instruct~\citep{bai2025qwen3vltechnicalreport}, and Qwen3-VL-30B-A3B-Instruct~\citep{bai2025qwen3vltechnicalreport}. The omni-modal family contains Qwen2.5-Omni-7B~\citep{Qwen25Omni} and Qwen3-Omni-30B-A3B-Instruct~\citep{Qwen3Omni}. The proprietary set contains Gemini-3-Flash and GPT-5-Mini. Official checkpoints or APIs are used with default inference settings unless stated otherwise.\footnote{\raggedright Hugging Face checkpoints: \texttt{OpenGVLab/InternVL3\_5-4B}, \texttt{OpenGVLab/InternVL3\_5-8B}, \texttt{Qwen/Qwen3-VL-8B-Instruct}, \texttt{Qwen/Qwen3-VL-30B-A3B-Instruct}, \texttt{Qwen/Qwen2.5-Omni-7B}, and \texttt{Qwen/Qwen3-Omni-30B-A3B-Instruct}; APIs: \texttt{gemini-3-flash-preview} and \texttt{gpt-5-mini}.}

In addition to the eight reference baselines, the proposed \textbf{TimeOmni-v} (Section~\ref{sec:TimeOmni-v}) is evaluated under the same protocol. Its multimodal LLM half is initialized from Qwen2.5-Omni-7B~\citep{Qwen25Omni} and its time-series encoder branch from Chronos-2~\citep{Chronos2}; the multimodal LLM is fine-tuned with LoRA~\citep{Hu2022LoRA} while the time-series branch, the modality adapter, and the forecast head are fine-tuned. The remaining training details and per-dataset hyperparameters are deferred to Appendix~\ref{app:TimeOmni-v-training}.

Three controlled baselines separate the contribution of the time-series integration from that of fine-tuning and of the LLM. The \emph{adapted TS-as-text baseline} fine-tunes the same Qwen2.5-Omni-7B backbone with the identical LoRA configuration and per-dataset budget as TimeOmni-v, with the time series passed as lossless inline text. The \emph{non-LLM baselines} reuse the feature extractors of TimeOmni-v: Chronos-2 for time series, the frozen Qwen2.5-Omni vision encoder under the same preprocessing for vision, and a BERT encoder~\citep{devlin2019bert} for text. On these features, single-modality linear probes and a tri-modal late-fusion classifier over the concatenated features are trained for classification, and an MLP regression head over the same concatenated features is trained for forecasting. Finally, the TimeOmni-v time-series integration (Chronos-2 encoder branch, modality-aware token layout, and scatter injection) is ported to InternVL3.5-8B, a backbone with a different vision stack, and evaluated on CoVLA, AgiBotWorld, MIMIC, and FinMultiTime.

\begin{table}[t]
\centering
\caption{Main results on MMTA. swF1 ($\uparrow$) for classification, swMAPE ($\downarrow$) for forecasting. Best per column in \textbf{bold}, second best \underline{underlined}. ``---'' marks the cell where the model produced no parsable answer on any sample (Qwen3-VL-30B-A3B-Inst.\ runs out of memory on every AgiBot sample); per-row averages aggregate only over defined cells.}
\label{tab:main_results}
\setlength{\tabcolsep}{4pt}
\resizebox{\linewidth}{!}{\begin{tabular}{l cccccc ccc cc}
\toprule
& \multicolumn{6}{c}{Classification (swF1 $\uparrow$)} & \multicolumn{3}{c}{Forecasting (swMAPE $\downarrow$)} & \multicolumn{2}{c}{Overall} \\
\cmidrule(lr){2-7} \cmidrule(lr){8-10} \cmidrule(lr){11-12}
Model & AgiBot & CoVLA & CUHK-X & FuFa & HoloA & MIMIC & PixelRec & SP500 & Terra & Avg-F1 $\uparrow$ & Avg-MAPE $\downarrow$ \\
\midrule
\multicolumn{12}{c}{\textit{Open-source vision-language MLLMs}} \\
\midrule
InternVL3.5-4B           & 0.034 & 0.495 & 0.218 & 0.049 & 0.340 & 0.528 & 150.0 & 4.77 & 318.4 & 0.277 & 157.7 \\
InternVL3.5-8B           & 0.057 & 0.514 & 0.253 & 0.050 & 0.299 & 0.565 & \underline{98.0}  & 5.44 & 156.1 & 0.290 & 86.5  \\
Qwen3-VL-8B-Instruct     & 0.108 & \underline{0.592} & 0.084 & 0.049 & 0.269 & \textbf{0.655} & 145.5 & 5.49 & 77.4  & 0.293 & 76.1  \\
Qwen3-VL-30B-A3B-Inst.   & ---   & 0.577 & 0.084 & 0.129 & 0.000 & 0.585 & 129.2 & 4.75 & \underline{75.6}  & 0.275 & 69.8  \\
\midrule
\multicolumn{12}{c}{\textit{Open-source omni-modal MLLMs}} \\
\midrule
Qwen2.5-Omni-7B          & 0.074 & 0.424 & 0.099 & 0.198 & 0.257 & 0.555 & 104.4 & 7.73 & 177.8 & 0.268 & 96.6  \\
Qwen3-Omni-30B-A3B-Inst. & 0.052 & 0.491 & 0.244 & 0.129 & 0.234 & \underline{0.642} & 98.0  & 6.72 & \textbf{65.4}  & 0.299 & \textbf{56.7}  \\
\midrule
\multicolumn{12}{c}{\textit{Proprietary models}} \\
\midrule
GPT-5-mini               & 0.116 & 0.586 & 0.251 & 0.093 & \underline{0.569} & 0.602 & \textbf{91.3}  & 4.94 & 104.0 & 0.370 & 66.7  \\
Gemini-3-flash           & \underline{0.126} & 0.560 & \textbf{0.463} & \underline{0.315} & 0.561 & 0.628 & 91.7  & \textbf{4.45} & 96.1  & \underline{0.442} & \underline{64.1}  \\
\midrule
TimeOmni-v (Ours)        & \textbf{0.844} & \textbf{0.759} & \underline{0.343} & \textbf{0.362} & \textbf{0.694} & 0.602 & 108.4 & \underline{4.45} & 159.6 & \textbf{0.601} & 90.8  \\
\bottomrule
\end{tabular}}
\end{table}

\subsection{Tasks and Metrics}
\label{sec:metrics}

MMTA contains two evaluation formats, each paired with one headline metric.

Generation models may occasionally fail to return a valid, parsable answer (\textit{e.g.}, when the serialized time series exceeds the model's context window or the model deviates from the requested output format).\footnote{All open-source baselines are evaluated with the standard \texttt{transformers} inference pipeline at the model's default context length and precision. When the tokenized prompt exceeds the model's context window, or when decoding triggers a CUDA out-of-memory error on the available hardware, the sample is recorded as an inference failure and counted against the success rate.} To penalize such failures alongside ordinary errors, the \emph{success rate} -- the fraction of samples that produce a parsable answer -- is folded into both headline metrics.

\paragraph{Classification: swF1.}
A classification task asks the model to select one option from a candidate set. The base metric is Macro-F1, computed by averaging per-class F1 scores (one-vs-rest) to balance performance across classes. Unparsable outputs are treated as misclassifications, so the headline metric is success-weighted Macro-F1, $\text{swF1} = \text{Macro-F1} \times \text{SuccessRate}$, which penalizes both class confusion and frequent parsing failures.

\paragraph{Forecasting: swMAPE.}
A forecasting task asks the model to output multi-step numerical predictions over a fixed horizon. The base metric is $\text{MAPE} = (100\%/n)\sum_{i=1}^{n}|(y_i-\hat{y}_i)/y_i|$, and the headline metric is success-weighted MAPE, $\text{swMAPE} = \text{MAPE}/\text{SuccessRate}$, which penalizes both large errors and frequent parsing failures.

\paragraph{Decomposed metrics.}
For every model and test set, Appendix~\ref{app:decomposed} also reports the base metric on the parsed subset, the success rate, and a failure taxonomy (CUDA OOM, other inference failures such as API refusals, and unparsable outputs). The modality-subset ablation and the controlled baselines are reported as relative changes in Macro-F1 for classification and in MAE for forecasting.

\section{Results and Discussion}
\label{sec:results}

\subsection{Main Results}
\label{sec:results-main}

\paragraph{Headline.}
TimeOmni-v attains the highest Avg-swF1 (0.601 versus 0.442 for the next model, Gemini-3-flash) and the largest single-column gains on AgiBot (0.844), CoVLA (0.759), and HoloAssist (0.694). On Avg-swMAPE TimeOmni-v is mid-pack (90.8); the three forecasting domains remain comparable across recent baselines, and TimeOmni-v does not lead them.

\paragraph{Where the gap concentrates.}
Results show that TimeOmni-v's classification advantage concentrates on long-video, long-TS samples, and is markedly smaller where one or both modalities are short or fixed-length. On the long subset (AgiBot, CoVLA, and HoloAssist), every baseline stays below 0.6 swF1, while TimeOmni-v reports the highest swF1 on each. Table~\ref{tab:dataset_stats} shows what these three share: each pairs multi-second to multi-minute video with hundreds-to-thousands of TS steps. AgiBot reaches 155.7\,s of egocentric video and 4{,}627-step state traces; HoloAssist spans 20--80\,s with 600--2{,}401-step sensor windows; CoVLA samples extend up to 30\,s. On the short-or-fixed subset, Gemini-3-flash leads on CUHK-X (sub-second clips) by 12.0\,pp, and TimeOmni-v leads on Future Factories (fixed 16-step TS) by 4.7\,pp, a margin within the uncertainty of its 46 test samples (Appendix~\ref{app:bootstrap}). A plausible reading of this split is that when the TS is short or fixed-length, its textual serialization fits inside the LLM context with room to spare, and a strong general visual backbone is enough to read it alongside the video. When the TS becomes long, serialization instead consumes the alignment budget and crowds out the co-occurring video frames; on the two longest-series domains, where TS-as-text fine-tuning runs out of memory (Section~\ref{sec:results-ablation}), a dedicated TS encoder is what makes training possible.

\paragraph{Scale does not close the long-TS gap.}
On AgiBot, the 30B-class Qwen3-VL-30B-A3B-Instruct runs out of memory on every sample, and every other baseline, including the 30B-class Qwen3-Omni-30B-A3B-Instruct and both proprietary models, stays under 0.13 swF1. Increasing parameter count alone does not move either family beyond the failure mode named in Section~\ref{sec:intro}: a long serialized time series consumes the alignment budget that the LLM would otherwise spend on the co-occurring video frames.

\subsection{Inference Efficiency on the Worst-Case TS Domain}
\label{sec:results-efficiency}

\begin{table}[t]
\centering
\caption{Inference cost of TimeOmni-v and its base model on AgiBot test (131 samples, the longest median TS in MMTA; single H200, max\_new\_tokens=8, identical dynamic-batching budgets). TS tokens count the time-series portion of the prompt only. \emph{Completed} counts samples that finish without CUDA OOM in this run. AgiBot is reported as a worst-case TS-length study; per-sample latency and VRAM scale with TS length, the token compression ratio does not.}
\label{tab:efficiency}
\setlength{\tabcolsep}{6pt}
\small
\begin{tabular}{l cccc}
\toprule
Model & TS tokens $\downarrow$ & Per sample (s) $\downarrow$ & Peak VRAM (GB) $\downarrow$ & Completed $\uparrow$ \\
\midrule
Qwen2.5-Omni-7B (TS-as-text) & 167{,}041 & 27.70 & 108.6 & 113 / 131 \\
TimeOmni-v                       & \textbf{722}     & \textbf{1.20}  & \textbf{35.0}  & \textbf{131 / 131} \\
\midrule
Ratio (baseline / TimeOmni-v)    & 232.8$\times$ & 23.0$\times$ & 3.1$\times$ & --- \\
\bottomrule
\end{tabular}
\end{table}

\paragraph{Token compression: dataset-independent.}
For an input time series of shape $T\!\times\!C$, TimeOmni-v represents the TS slice as $P\!\cdot\!C+2$ tokens, where $P=\lceil T/p\rceil$ is the number of Chronos-2 patches at patch size $p\!=\!16$ and the $+2$ are boundary tokens. Text serialization, by contrast, produces roughly $T\!\cdot\!C\!\cdot\!k$ sub-word tokens, where $k$ counts the sub-word tokens spent on each numeric value together with the surrounding formatting (timestamps, channel names, separators, decimal digits) and is typically larger than 10 in practice. The compression ratio is therefore $TCk/(TC/p+2)\approx p\!\cdot\!k$ in the large-$T,C$ limit, fixed by the patch size and the serialization verbosity rather than by dataset content. On AgiBot the median TS portion drops from 167{,}041 sub-word tokens to 722 ($232.8\times$), consistent with $p\!=\!16$ and an effective $k\!\approx\!14.5$.

\paragraph{Per-sample speedup: dataset-dependent.}
Token compression and end-to-end speedup are not the same factor. On AgiBot per-sample latency drops 23.0$\times$ and peak VRAM 3.1$\times$. Video tokens, system prompt, and decoding overhead are unchanged; only the TS slice shrinks, and its share of the total prompt determines how much of the cost actually moves. The implication is that this speedup is largest on long-TS domains (AgiBot is the worst case) and shrinks toward unity on short-TS image$+$TS domains.

\paragraph{Reliability.}
Eighteen of the 131 AgiBot samples (14\%) cause Qwen2.5-Omni-7B to exceed the 140~GB H200 activation budget in this run, whereas TimeOmni-v finishes all 131 samples. The reliability gap shares its origin with the swF1 gap: a long serialized TS is not only expensive but, on a non-trivial fraction of samples, infeasible at all.

\subsection{Ablations and Controlled Comparisons}
\label{sec:results-ablation}

\begin{figure}[t]
\centering
\includegraphics[width=\linewidth]{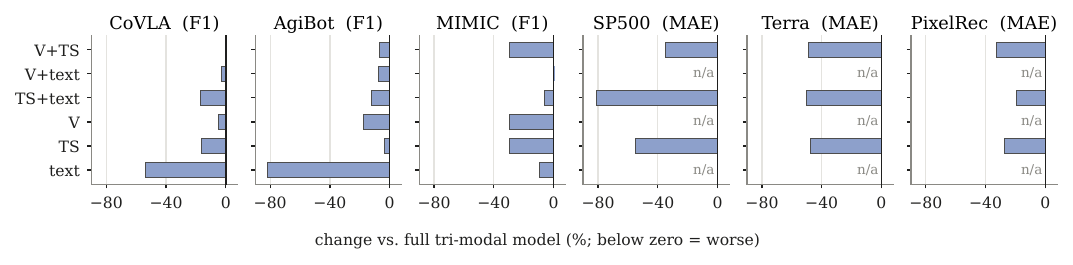}
\caption{Training-time modality ablation. Each bar is TimeOmni-v trained and tested with the kept modalities only (V = vision), as the relative change (\%) against the full tri-modal model in Macro-F1 for classification and in MAE for forecasting; MAE changes are sign-flipped so that a bar below zero is always worse. ``n/a'': forecasting without time series is undefined, since the forecast head requires time-series input.}
\label{fig:modality}
\end{figure}

Figure~\ref{fig:modality} retrains TimeOmni-v on every single- and two-modality subset and tests it on the same subset, so each change measures what the removed modalities contribute; removing a modality only at test time measures reliance instead and is reported for all nine domains in Appendix~\ref{app:inference-ablation}. Figure~\ref{fig:controls} compares TimeOmni-v with the controlled baselines of Section~\ref{sec:evaluated-models}.

\begin{figure}[t]
\centering
\includegraphics[width=\linewidth]{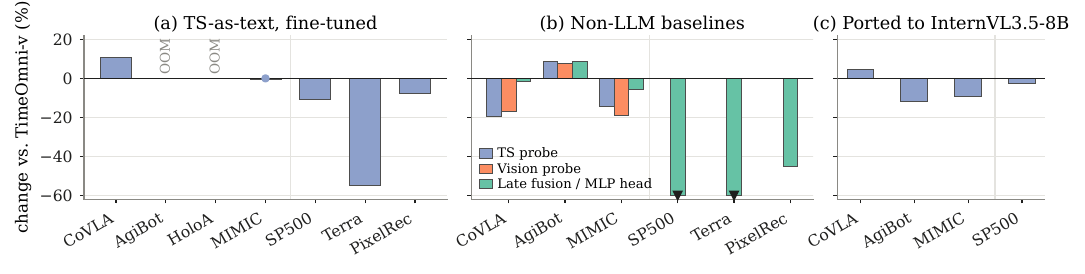}
\caption{Controlled comparisons, as relative changes (\%) in Macro-F1 for classification and in MAE for forecasting (sign-flipped, so a bar below zero is always worse than the reference). (a) Qwen2.5-Omni-7B fine-tuned with time series as text under the same LoRA configuration and budget, against TimeOmni-v; OOM: training runs out of memory on an H200 140\,GB GPU at batch size one. (b) Non-LLM baselines: linear probes and late fusion on the frozen TimeOmni-v features, against TimeOmni-v; triangles mark bars that extend beyond the axis. (c) The TimeOmni-v integration ported to InternVL3.5-8B, against the Qwen2.5-Omni-7B version (swF1 and swMAPE; both parse every sample).}
\label{fig:controls}
\end{figure}

\paragraph{The dominant modality differs across datasets, but every reduced input loses somewhere.}
No subset improves on the full model by more than 0.7\%, and each single- or two-modality subset is worse on at least two of the datasets where it is defined. Where the losses vanish, they locate the dominant modality. Dropping text costs nothing on CoVLA, whose text only states the task and the channels, so CoVLA is vision-led; dropping time series costs nothing on MIMIC, where the report and the X-ray carry more of the prognosis signal than the vital signs, so MIMIC is text-led. On AgiBot every subset loses, and on the three forecasting datasets every subset raises MAE by 19.7--81.4\%. The time series alone trails the full model by 16.4\% on CoVLA, so the partial overlap between its rule-derived labels and the input channels (Section~\ref{sec:bench-video}) does not make the task solvable from the series.

\paragraph{Fine-tuning explains much of the short-series gap; the encoder makes long series trainable.}
The TS-as-text baseline (Figure~\ref{fig:controls}a) shares the backbone, the LoRA configuration, and the training budget with TimeOmni-v, so it isolates how the time series enters the model. On MIMIC it matches TimeOmni-v ($-$0.4\%), and on CoVLA it is ahead ($+$11.0\%), where the rule-derived labels reward exact numeric readout, which lossless text keeps and patch compression does not. On AgiBot and HoloAssist the textualized prompts run out of memory, so the encoder is what makes fine-tuning possible. On forecasting the TS-as-text baseline has 7.8--55.0\% higher MAE, a gap that also includes the change from digit decoding to the forecast head.

\paragraph{Non-LLM baselines solve some classification tasks but fail at forecasting.}
On AgiBot, the TS probe, the vision probe, and late fusion exceed TimeOmni-v by 7.9--8.9\% (Figure~\ref{fig:controls}b), whereas on CoVLA and MIMIC they trail it by 1.7--19.6\%. The AgiBot labels are therefore learnable without an LLM, and the zero-shot MLLMs in Table~\ref{tab:main_results} stay far below what a linear probe extracts from the same inputs. The same tri-modal features with an MLP regression head collapse on forecasting, raising MAE by 45.1\% on PixelRec and by more than an order of magnitude on SP500 and Terra.

\paragraph{The time-series integration is not tied to one backbone.}
Ported to InternVL3.5-8B, a backbone with a different vision stack, the integration (Figure~\ref{fig:controls}c) parses every sample and stays within 12\% of the Qwen2.5-Omni-7B version on all four tested datasets, ahead on CoVLA and behind on AgiBot and MIMIC. On AgiBot its 0.746 swF1 still exceeds every zero-shot baseline in Table~\ref{tab:main_results} by more than 0.6.

\section{Conclusion}

We introduce \textbf{MMTA}, a collection of tri-modal tasks for multimodal temporal analysis with diverse cross-modal complementarity patterns, spanning nine real-world domains, two cross-modal layouts (video$+$TS and image$+$TS), and a 3{,}524-sample test split that covers both classification and forecasting. We evaluate eight representative open-source and proprietary MLLMs on MMTA. Two key insights emerge. First, the dominant modality differs across domains, yet every single- and two-modality variant of the same model falls behind the tri-modal one on part of the benchmark. Second, the evaluated MLLMs share a failure mode in which serializing time series as text inflates the prompt, causes out-of-memory failures and unparsable outputs on long sequences, and leaves accuracy low even when every sample is answered; neither model scale nor proprietary tuning closes this long-sequence gap. To this end, we present \textbf{TimeOmni-v}, a framework that integrates time series into multimodal LLMs as a first-class modality while remaining compatible with the existing visual and language pretraining. On MMTA, TimeOmni-v lifts Avg-swF1 over the strongest baseline and can be fine-tuned on long-sequence domains where the same backbone with textualized time series runs out of memory. By providing a tri-modal benchmark and a preliminary integration recipe, this work takes a step toward Multimodal LLMs that read time series as faithfully as they already read vision and language, enabling end-to-end decision-making over real-world dynamical systems where evidence is distributed across modalities.

\begin{ack}
This work was carried out with internal resources of Tsinghua University and the Shanghai Artificial Intelligence Laboratory and received no third-party funding. The authors declare no competing interests.
\end{ack}

\bibliographystyle{unsrtnat}
\bibliography{mmta}

\newpage
\appendix

\section{Real-World Multimodal Temporal Datasets in MMTA}
\label{app:realworld_datasets}

This appendix complements Section~\ref{sec:benchmark} with the dataset-level details that do not fit into the main text. For each of the nine MMTA domains, the corresponding subsection describes (i) the source corpus and the modalities it supplies, (ii) the construction pipeline that turns raw releases into one MMTA instance under the unified schema, and (iii) a representative prompt example that shows what the model actually sees at evaluation time. Three modalities recur across the corpus: \textbf{time series} (physiological signals, joint and gripper traces, vehicle states, IMU streams, gaze and head trajectories, daily interaction counts, stock prices, climate variables); \textbf{vision} (egocentric RGB clips, dashcam segments, skeleton renders, on-site surveillance video, chest X-rays, item thumbnails, candlestick charts, remote-sensing imagery); and \textbf{text} (task instructions, fine-grained action sequences, cycle reports, radiology reports, item metadata, news snippets, geographic descriptions). Domains are grouped by cross-modal layout: five \emph{video$+$TS} domains first, then four \emph{image$+$TS} domains. Standardized statistics over the 3{,}524-sample MMTA test split are summarized in Table~\ref{tab:dataset_stats}.

\begin{table*}[t]
\centering
\footnotesize
\setlength{\tabcolsep}{4pt}
\renewcommand{\arraystretch}{1.05}
\caption{Standardized dataset statistics of MMTA. \emph{Layout} indicates the cross-modal layout (V$+$TS = video$+$TS; I$+$TS = image$+$TS). \emph{Train} and \emph{Test} are the split sizes. \emph{Duration} is the per-sample video length in seconds (min--max), reported only for V$+$TS domains. \emph{Seq.\,Len.} is the time-series length per sample, \emph{Dim.} is the number of time-series channels, and \emph{Visual} summarizes the per-sample visual content. The MIMIC entry merges the mortality and discharge subtasks (75 test samples each).}
\resizebox{\linewidth}{!}{\begin{tabular}{@{}llccccccl@{}}
\toprule
\textbf{Dataset} & \textbf{Layout} & \textbf{Train} & \textbf{Test} & \textbf{Duration (s)} & \textbf{Seq.\,Len.} & \textbf{Dim.} & \textbf{Visual} & \textbf{Task} \\
\midrule
AgiBotWorld      & V$+$TS & 1{,}177 & 131   & 7.9--155.7   & 194--4{,}627 & 10 & egocentric clip            & Classification (4-way) \\
CoVLA            & V$+$TS & 13{,}437 & 1{,}344 & 0.3--30.0  & 6--600       & 5  & dashcam clip               & Classification (3-way) \\
CUHK-X           & V$+$TS & 2{,}629 & 308   & 0.05--16.05  & 1--816       & 7  & skeleton clip              & Classification (40-way) \\
Future Factories & V$+$TS & 49 & 46    & 19.5--69.9   & 16 (fixed)   & 36 & paired-camera video        & Classification (4-way) \\
HoloAssist       & V$+$TS & 4{,}991 & 945   & 20.0--80.4   & 600--2{,}401 & 9  & egocentric clip            & Classification (binary) \\
MIMIC            & I$+$TS & 150 & 150   & ---          & 4--39        & 6  & 1--5 chest X-rays          & Classification (binary) \\
PixelRec         & I$+$TS & 931 & 200   & ---          & 14--38       & 1  & 1 cover image              & Forecasting \\
FinMultiTime     & I$+$TS & 3{,}368 & 200   & ---          & 82--124      & 1  & 1 candlestick chart        & Forecasting \\
Terra            & I$+$TS & 2{,}000 & 200   & ---          & 100 (fixed)  & 7  & 4--7 images (satellite and map layers) & Forecasting \\
\bottomrule
\end{tabular}}
\label{tab:dataset_stats}
\end{table*}

\subsection{Robotics: AgiBotWorld}

\paragraph{Source and modalities.}
AgiBotWorld2026~\citep{AgiBotWorld2026} is a robotics dataset for embodied AI and imitation learning, focused on real-world robotic manipulation. The MMTA subset is built from selected task groups with complete episode metadata and synchronized visual/state streams. Each episode contains an explicit task instruction, a stepwise execution trace, and synchronized state and action trajectories. The visual modality is the top-head camera clip (\texttt{observation.images.top\_head}) over the full Task Frame, providing context around each error moment. The temporal modality contains 10 channels extracted from \texttt{observation.state}: six arm-position channels (\texttt{arm\_pos\_0}--\texttt{arm\_pos\_5}) and four end-effector velocity channels (\texttt{end\_vel\_0}--\texttt{end\_vel\_3}), with timestamps and frame indices retained as alignment metadata. The text modality combines the task instruction with the ordered fine-grained step sequence.
\textbf{Data source:} \url{https://huggingface.co/datasets/agibot-world/AgiBotWorld2026}.

\paragraph{Atomic instance and label.}
An atomic instance is one Task Frame within an episode. The model classifies the dominant anomaly cause among four labels: \emph{paused} (A), \emph{meaningless operation} (B), \emph{fall} (C), or \emph{other} (D). Task Frames containing multiple anomaly types are dropped to keep the label space single-valued. The final export contains 1{,}308 instances (1{,}177 train / 131 test).

\paragraph{Processing pipeline.}
(1) Read structured episode metadata for each usable segment and collect task-frame and instruction annotations.
(2) Identify Task Frame candidates from annotations whose frame type indicates a Task Frame.
(3) Collect error frames whose midpoints fall inside each Task Frame and map raw error descriptions onto the four anomaly categories.
(4) Load the corresponding episode trajectory, slice the state tensor over the Task Frame interval, and align the 10-channel state stream with the cropped video clip on the same timeline.
(5) Assemble the prompt from the task instruction, ordered step sequence, cropped video, and aligned state traces.

\begin{tcolorbox}[title={AgiBotWorld (concrete instance)},breakable]
\textbf{task\_type:} QA \quad \textbf{video:} top-head egocentric clip, $640\times 400$, 30 fps, 16.67\,s, 500 frames. \quad \textbf{ts\_shape:} $452\times 10$. \\[0.4em]
\begin{center}
  \includegraphics[width=0.8\linewidth]{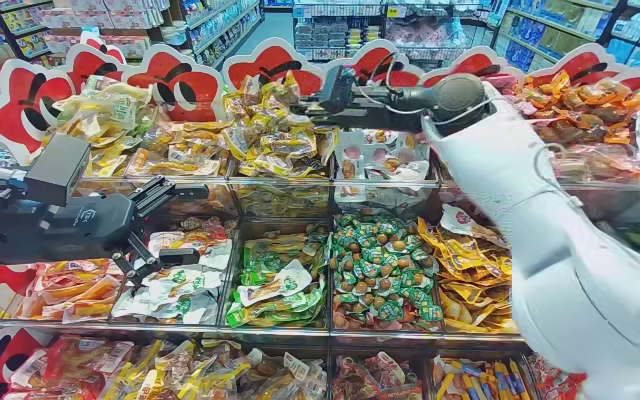}
\end{center}
\textbf{scene description:}
\begin{tcolorbox}[colback=white, colframe=black!20, boxrule=0.5pt, arc=2pt]
\footnotesize
This clip is from AgiBot World in a real retail shelf-rearrangement scene. The robot performs dual-arm manipulation near shelves and shopping carts.
\end{tcolorbox}
\textbf{text (task instruction $+$ step sequence):}
\begin{tcolorbox}[colback=white, colframe=black!20, boxrule=0.5pt, arc=2pt]
\footnotesize
Task instruction: \emph{Adjust the orange sausage set to the correct position on the shelf.} \\
Step sequence: \\
1. {[Straighten]} Adjust products. \\
2. {[Other]} Adjust the position of the orange sausage set with the left arm.
\end{tcolorbox}
\textbf{time series (10-channel state, format \texttt{timestamp: arm\_pos\_0=..., ..., end\_vel\_3=...}):}
\begin{tcolorbox}[colback=white, colframe=black!20, boxrule=0.5pt, arc=2pt]
\footnotesize\raggedright
12.400: arm\_pos\_0=0.356311, arm\_pos\_1=$-$0.029625, arm\_pos\_2=0.378649, \dots, end\_vel\_0=0.053887, end\_vel\_1=0.416501, end\_vel\_2=$-$0.360235, end\_vel\_3=1.185560 \\
12.433: arm\_pos\_0=0.358105, arm\_pos\_1=$-$0.018230, arm\_pos\_2=0.391642, \dots, end\_vel\_0=0.057594, end\_vel\_1=0.436849, end\_vel\_2=$-$0.387070, end\_vel\_3=1.130598 \\
\dots \\
27.433: arm\_pos\_0=0.381310, arm\_pos\_1=$-$0.083640, arm\_pos\_2=0.318216, \dots, end\_vel\_0=0.000105, end\_vel\_1=$-$0.000011, end\_vel\_2=$-$0.000281, end\_vel\_3=$-$0.000038
\end{tcolorbox}
\textbf{prompt (abridged):}
\begin{lstlisting}[basicstyle=\ttfamily\footnotesize,breaklines=true,escapeinside={(*}{*)},breakindent=0pt,columns=fullflexible,breakatwhitespace=true]
You are a robot video anomaly-reason classification assistant.

<scene_description> (*\textbf{scene description}*) </scene_description>

<video_capture>
The video is captured by the robot top-head camera
(observation.images.top_head), approximately 30 fps. Important: this
clip contains a complete task-frame segment, not only the short error
moment.
</video_capture>

<task_instruction> (*\textbf{instruction}*) </task_instruction>
<step_sequence>   (*\textbf{steps}*)       </step_sequence>

<error_definitions>
- A paused: operation is paused or stalled
- B meaningless operation: ineffective or redundant operation behavior
- C fall: object drops or a falling-related error occurs
- D other: other error causes not covered above
</error_definitions>

<numeric_field_explanations>
arm_pos_{0..5}: end-effector / arm position channels;
end_vel_{0..3}: end-effector velocity sub-channels.
</numeric_field_explanations>

<timeseries> (*\textbf{time series}*) </timeseries>
\end{lstlisting}
\textbf{question:} Based on the video, task instruction, step sequence, and timeseries above, what is the most appropriate anomaly cause category for this task-frame segment? \\
\textbf{options:} A. paused. \quad B. meaningless operation. \quad C. fall. \quad D. other. \\
\textbf{expected answer:} A.
\end{tcolorbox}

\subsection{Autonomous Driving: CoVLA}

\paragraph{Source and modalities.}
The CoVLA Dataset~\citep{CoVLA2024} consists of driving videos recorded on Japanese public roads, together with annotations generated with the assistance of VideoLLaMA~2. In the MMTA processing corpus, 6{,}010 driving sequences provide complete \texttt{states} and \texttt{front\_car} annotations for segment extraction. The visual stream is a temporally cropped braking segment from the original dashcam clip, focused on the event window where the braking decision occurs. The temporal stream contains five synchronized driving variables: \texttt{vEgo}, \texttt{aEgo}, \texttt{steeringAngleDeg}, \texttt{lead\_speed\_kmh}, and \texttt{lead\_a}. The upstream brake and gas signals are used to locate the braking interval but are not exposed as model inputs. The text modality consists of the scenario description and a short scene caption.
\textbf{Data source:} \url{https://huggingface.co/datasets/turing-motors/CoVLA-Dataset}.

\paragraph{Atomic instance and label.}
An atomic instance is one braking segment, defined as a continuous interval where \texttt{brakePressed} stays true for at least six frames. The model performs three-way brake-reason classification: \emph{lead-vehicle deceleration} (A), \emph{ego-vehicle turning} (B), or \emph{other reasons} (C). The final export contains 14{,}781 instances (13{,}437 train / 1{,}344 test).

\paragraph{Processing pipeline.}
(1) Read the per-frame vehicle-state and front-vehicle annotations.
(2) Detect continuous segments where \texttt{brakePressed} remains true for at least six frames.
(3) Assign the brake-reason label using synchronized evidence: \texttt{A} if the lead vehicle shows sustained deceleration; \texttt{B} if the ego vehicle exhibits turning behavior; \texttt{C} otherwise.
(4) Slice the aligned per-frame signals to the braking window, convert timestamps to seconds relative to the video start, and assemble the prompt from the cropped clip and the five-channel state stream.

\begin{tcolorbox}[title={CoVLA (concrete instance)},breakable]
\textbf{task\_type:} QA \quad \textbf{video:} dashcam clip, $1928\times 1208$, 20 fps, 4.65\,s, 93 frames. \quad \textbf{ts\_shape:} $93\times 5$. \\[0.4em]
\begin{center}
  \includegraphics[width=0.8\linewidth]{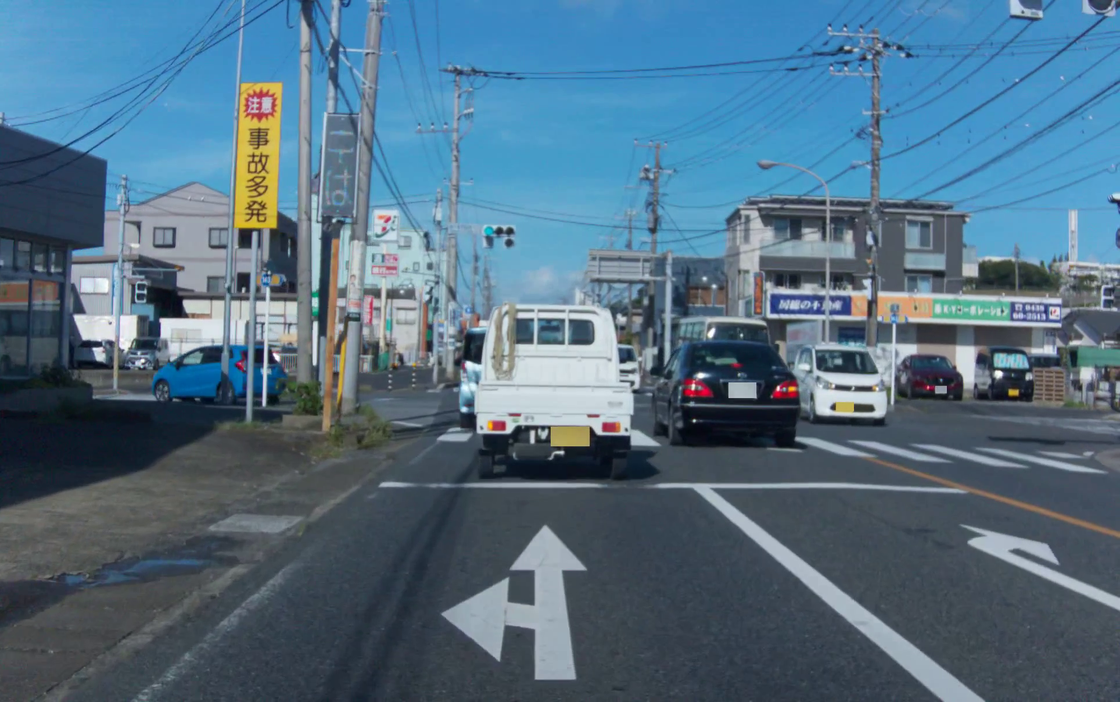}
\end{center}
\textbf{variable definitions:}
\begin{tcolorbox}[colback=white, colframe=black!20, boxrule=0.5pt, arc=2pt]
\footnotesize
\texttt{vEgo}: ego speed (m/s). \quad \texttt{aEgo}: longitudinal acceleration (m/s$^2$, negative $\Rightarrow$ deceleration). \quad \texttt{steeringAngleDeg}: steering wheel angle in degrees (large $|\cdot|$ $\Rightarrow$ turning). \quad \texttt{lead\_speed\_kmh}: leading-vehicle speed (km/h, \texttt{null} when no lead detected). \quad \texttt{lead\_a}: leading-vehicle acceleration (m/s$^2$, \texttt{null} when no lead detected).
\end{tcolorbox}
\textbf{time series (per-frame, braking onset $\to$ offset, format \texttt{t(s): vEgo=, aEgo=, steeringAngleDeg=, lead\_speed\_kmh=, lead\_a=}):}
\begin{tcolorbox}[colback=white, colframe=black!20, boxrule=0.5pt, arc=2pt]
\footnotesize\raggedright
25.5: vEgo=6.517, aEgo=0.197, steeringAngleDeg=$-$16.5, lead\_speed\_kmh=null, lead\_a=null \\
25.5: vEgo=6.505, aEgo=0.149, steeringAngleDeg=$-$17.6, lead\_speed\_kmh=null, lead\_a=null \\
\dots \\
30.0: vEgo=1.561, aEgo=$-$0.691, steeringAngleDeg=$-$2.0, lead\_speed\_kmh=null, lead\_a=null \\
30.0: vEgo=1.578, aEgo=$-$0.505, steeringAngleDeg=15.2, lead\_speed\_kmh=null, lead\_a=null
\end{tcolorbox}
\textbf{prompt (abridged):}
\begin{lstlisting}[basicstyle=\ttfamily\footnotesize,breaklines=true,escapeinside={(*}{*)},breakindent=0pt,columns=fullflexible,breakatwhitespace=true]
You are given a question-answering (QA) task for autonomous driving.
You are provided with a dashcam video segment from an autonomous driving
system (ego vehicle) traveling on public roads. This video is already
temporally cropped to only the braking event, and the per-frame sensor
data below is aligned to this same braking segment. Your task is to
determine the most likely reason for the sudden braking in this segment.
Do NOT output any analysis. Return only the option letter.

<video>
This video clip contains only the braking segment. The braking segment
contains 93 frames. No pre-brake or post-brake frames are included; the
sensor data below also contains only this exact braking segment.
</video>

<variables> (*\textbf{variable definitions}*) </variables>
<timeseries> (*\textbf{time series}*) </timeseries>
\end{lstlisting}
\textbf{question:} Based on the video content and the sensor data above, why did the ego vehicle suddenly brake during the segment? \\
\textbf{options:} A. The leading vehicle suddenly decelerated. \quad B. The ego vehicle was making a turn. \quad C. Other reason (e.g., pedestrian about to cross, traffic light turning red, entering a tunnel). \\
\textbf{expected answer:} C.
\end{tcolorbox}

\subsection{Human Action: CUHK-X}

\paragraph{Source and modalities.}
CUHK-X~\citep{CUHKX2024} is a large-scale multimodal dataset for human action recognition, understanding, and reasoning, containing 64{,}342 samples collected from 30 participants across diverse environments. It supplies seven synchronized modalities including RGB, infrared, depth, thermal, IMU, and other sensor streams. The MMTA subset uses CUHK-S/HAR and constructs a pose- and inertial-sensor based action classification task. The visual stream is a synthetic skeleton-rendered clip from the HAR release, so body-pose evolution over time is encoded directly in the video. The temporal stream consists of five wearable IMU streams mapped to a common device space (\texttt{WTC}, \texttt{WTLA}, \texttt{WTLL}, \texttt{WTRA}, \texttt{WTRL}); each step contains \texttt{time\_s}, \texttt{device\_id}, three accelerometer channels (\texttt{acc\_x\_g}, \texttt{acc\_y\_g}, \texttt{acc\_z\_g}), and three gyroscope channels (\texttt{gyro\_x\_dps}, \texttt{gyro\_y\_dps}, \texttt{gyro\_z\_dps}).
\textbf{Data source:} \url{https://github.com/openaiotlab/CUHK-X}.

\paragraph{Atomic instance and label.}
An atomic instance is one action trial. The model classifies the action over 40 categories (\texttt{action\_id} 0--39). The final export contains 2{,}937 instances (2{,}629 train / 308 test).

\paragraph{Processing pipeline.}
(1) Match each skeleton-rendered clip with the corresponding action trial and IMU records using the dataset-provided hierarchy.
(2) Merge the available IMU streams in each trial, clean column names, and sort rows by timestamp.
(3) Normalize device names and derive a stable \texttt{device\_id} mapping.
(4) Downsample each second independently for each device, keeping up to five representative samples per second.
(5) Use relative time from the first sampled IMU record and align the resulting sequence with the corresponding skeleton clip.

\begin{tcolorbox}[title={CUHK-X (concrete instance)},breakable]
\textbf{task\_type:} action classification (40-way) \quad \textbf{video:} skeleton clip, $640\times 640$, 20 fps, 0.75\,s, 15 frames. \quad \textbf{ts\_shape:} $43\times 7$. \\[0.4em]
\begin{center}
  \includegraphics[width=0.8\linewidth]{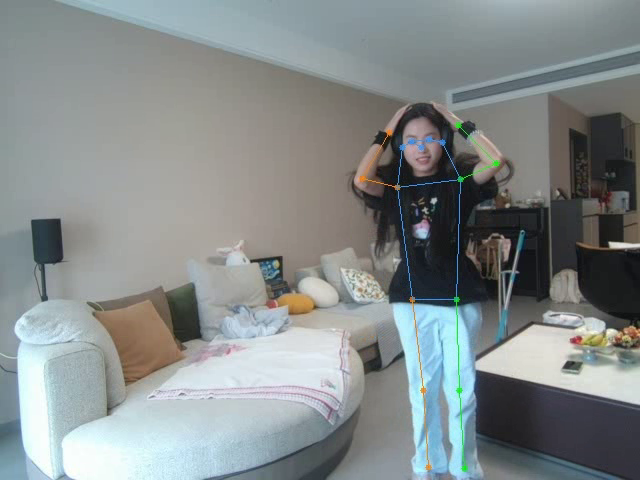}
\end{center}
\textbf{device definitions:}
\begin{tcolorbox}[colback=white, colframe=black!20, boxrule=0.5pt, arc=2pt]
\footnotesize
\texttt{device\_id=0}: WTC (chest). \quad \texttt{device\_id=1}: WTLA (left arm). \quad \texttt{device\_id=2}: WTLL (left leg). \quad \texttt{device\_id=3}: WTRA (right arm). \quad \texttt{device\_id=4}: WTRL (right leg).
\end{tcolorbox}
\textbf{action space (excerpt):}
\begin{tcolorbox}[colback=white, colframe=black!20, boxrule=0.5pt, arc=2pt]
\footnotesize
0: Wash\_face, \quad 1: Brush\_teeth, \quad 2: Comb\_hair, \quad \ldots, \quad 28: Jog\_in\_place, \quad \textbf{29: Do\_squats}, \quad 30: Do\_jumping\_jacks, \quad \ldots, \quad 39: Take\_body\_temperature.
\end{tcolorbox}
\textbf{time series (one row per device per timestamp, format \texttt{time\_s: device\_id=N, acc\_x\_g=, acc\_y\_g=, acc\_z\_g=, gyro\_x\_dps=, gyro\_y\_dps=, gyro\_z\_dps=}):}
\begin{tcolorbox}[colback=white, colframe=black!20, boxrule=0.5pt, arc=2pt]
\footnotesize\raggedright
0.000: device\_id=4, acc\_x\_g=$-$0.31, acc\_y\_g=0.63, acc\_z\_g=$-$0.75, gyro\_x\_dps=116, gyro\_y\_dps=90, gyro\_z\_dps=$-$4 \\
0.004: device\_id=1, acc\_x\_g=$-$0.13, acc\_y\_g=0.64, acc\_z\_g=0.72, gyro\_x\_dps=5, gyro\_y\_dps=112, gyro\_z\_dps=29 \\
0.041: device\_id=3, acc\_x\_g=$-$0.17, acc\_y\_g=$-$0.53, acc\_z\_g=0.83, gyro\_x\_dps=11, gyro\_y\_dps=34, gyro\_z\_dps=$-$2 \\
\dots \\
1.380: device\_id=1, acc\_x\_g=$-$0.58, acc\_y\_g=$-$0.08, acc\_z\_g=$-$0.11, gyro\_x\_dps=$-$3, gyro\_y\_dps=59, gyro\_z\_dps=$-$11
\end{tcolorbox}
\textbf{prompt (abridged):}
\begin{lstlisting}[basicstyle=\ttfamily\footnotesize,breaklines=true,escapeinside={(*}{*)},breakindent=0pt,columns=fullflexible,breakatwhitespace=true]
You are solving an action-classification task for CUHK-X multimodal HAR.
The video segment and downsampled IMU samples belong to the same action
instance.

<field_definitions>
time_s: relative timestamp (s) from the first sampled IMU record;
device_id: numeric device identifier (see <device_definitions>);
acc_{x,y,z}_g: acceleration (g);
gyro_{x,y,z}_dps: angular velocity (deg/s).
</field_definitions>

<device_definitions> (*\textbf{device defs}*) </device_definitions>
<action_space>      (*\textbf{40 action labels}*) </action_space>
<timeseries>        (*\textbf{time series}*)      </timeseries>
\end{lstlisting}
\textbf{question:} Which action category (action\_id) best matches this segment? Return only one integer action\_id from the action\_space. \\
\textbf{expected answer:} \texttt{29} (\emph{Do\_squats}).
\end{tcolorbox}

\subsection{Manufacturing: Future Factories}

\paragraph{Source and modalities.}
The Future Factories Platform~\citep{Harik2024FutureFactories} is a multimodal dataset family for industrial monitoring, capturing manufacturing production cycles with synchronized PLC sensor streams, on-site camera footage, and textual cycle reports. The MMTA subset is collected from an industry-grade assembly line at the Future Factories Lab, University of South Carolina; the processed source stream is grouped into 95 production cycles. The physical process performs collaborative assembly of a Lego-style modular rocket comprising four components: a base, two body segments (Body1, Body2), and a nose cone. Robot R01 introduces the tray; R02 retrieves the body segments; R03 mounts them onto the base before adding the nose cone; R04 then performs disassembly to reset the cycle. The visual stream is the on-site surveillance footage captured by two synchronized cameras monitoring the assembly station. The temporal stream contains 36 PLC channels: 12 gripper-related signals and 24 robot joint-angle channels. The text modality provides a cycle report with process overview, signal definitions, and time range.
\textbf{Data source:} \url{https://www.kaggle.com/datasets/ramyharik/ff-2024-08-13-multi-modal-dataset-33}.

\paragraph{Atomic instance and label.}
An atomic instance is one production cycle. Defects are introduced by manually removing required parts before the cycle starts; the model classifies the cycle into one of four labels:
\emph{Correct} (A; 10 samples, 10.53\%),
\emph{NoNose} (B; 33, 34.74\%),
\emph{NoNose,NoBody2} (C; 11, 11.58\%),
\emph{NoNose,NoBody2,NoBody1} (D; 41, 43.16\%).
The final export contains 95 instances (49 train / 46 test).

\paragraph{Processing pipeline.}
(1) Parse the timestamped source records and sort them by timestamp.
(2) Segment the stream into 95 production cycles using the cycle delimiters in the source.
(3) Downsample each cycle by keeping the last record in every 2-second bucket.
(4) Cut the two synchronized camera recordings to the cycle interval and stitch the two views side by side into one video.
(5) Sample 16 timestamps uniformly across the cycle, align the 36-channel PLC trace to them, and build the prompt from the cycle report text, the cycle video, and the aligned time series.

\begin{tcolorbox}[title={Future Factories Platform},breakable]
\textbf{task\_type:} QA \\[0.2em]
\textbf{video:} paired-camera surveillance video of the full cycle, with the two views stitched side by side, $2160\times 720$, 10 fps.
\begin{center}
  \includegraphics[width=0.8\linewidth]{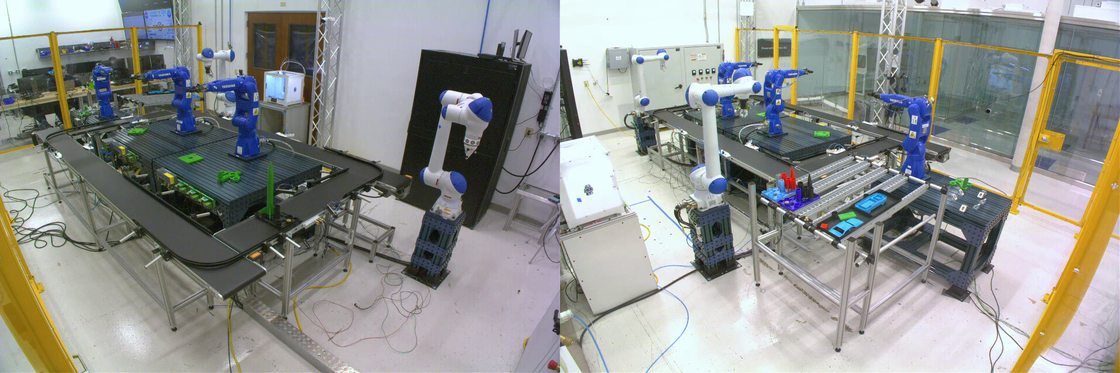}
\end{center}
\textbf{text:}
\begin{tcolorbox}[colback=white, colframe=black!20, boxrule=0.5pt, arc=2pt]
\footnotesize
Time range (UTC): xxx \quad Cycle ID: xxx \quad Records: xxx \\
Process overview: xxx \\
Selected signals and meanings: xxx \\
Video description (cycle-specific multi-camera recordings): xxx
\end{tcolorbox}
\textbf{time series (16 sampled timestamps, 36 channels):}
\begin{tcolorbox}[colback=white, colframe=black!20, boxrule=0.5pt, arc=2pt]
\footnotesize\raggedright
timestamp: \\
I\_R0X\_Gripper\_Pot=\dots, I\_R0X\_Gripper\_Load=\dots, M\_R0X\_SJointAngle\_Degree=\dots, M\_R0X\_LJointAngle\_Degree=\dots, M\_R0X\_UJointAngle\_Degree=\dots, M\_R0X\_RJointAngle\_Degree=\dots, M\_R0X\_BJointAngle\_Degree=\dots, M\_R0X\_TJointAngle\_Degree=\dots \\
......
\end{tcolorbox}
\textbf{prompt:}
\begin{lstlisting}[basicstyle=\ttfamily\footnotesize,breaklines=true,escapeinside={(*}{*)},breakindent=0pt,columns=fullflexible,breakatwhitespace=true]
You are given a question-answering (QA) task. Scenario: this cycle
comes from the Future Factories testbed. A tray is introduced at the
material handling station (MHS), transported by a closed-loop conveyor
between workcells, assembled by robots (R01-R03), and later disassembled
by R04 before the tray returns to MHS. The goal is to judge whether the
cycle proceeds normally or if a missing-part anomaly occurs. Use the
video(s), cycle report text, and time-series sensor signals as
evidence to answer the multiple-choice question. Return only the
option letter.

Note: the timestamps and time-series rows below are aligned to the
sampled timestamps in the cycle (one row per sampled timestamp,
time-ordered).

<report>
Cycle report text:
(*\textbf{text}*)
</report>

<timeseries>
Aligned time series (each line is one timestamp; missing values empty):
(*\textbf{time series}*)
</timeseries>
\end{lstlisting}
\textbf{question:} Did anything go wrong during this production cycle? Select the single best option that describes the observed issue (or choose \emph{correct} if no issue is present). \\
\textbf{options:} A. correct \quad B. NoNose \quad C. NoNose,NoBody2 \quad D. NoNose,NoBody2,NoBody1
\end{tcolorbox}

\subsection{AR Assistance: HoloAssist}

\paragraph{Source and modalities.}
HoloAssist~\citep{HoloAssist2023} is a large-scale egocentric human interaction dataset for studying interactive AI assistants on real-world physical tasks. The annotation dump used in MMTA processing contains 1{,}758 videos, 12{,}542 coarse-grained action annotations, and 149{,}253 fine-grained action annotations. The dataset is suitable for temporal benchmarking because each coarse action may contain multiple fine-grained actions, some of which are annotated as incorrect and may later be corrected by the instructor or the performer. The visual stream is the egocentric RGB clip cropped to one coarse-grained action segment, so unrelated pre- and post-segment content is excluded. The temporal stream is built from synchronized HoloLens streams aligned to the RGB timeline: gaze direction (\texttt{gaze\_dir\_x}, \texttt{gaze\_dir\_y}, \texttt{gaze\_dir\_z}), head translation (\texttt{tx}, \texttt{ty}, \texttt{tz}), and gyroscope measurements (\texttt{gyro\_x}, \texttt{gyro\_y}, \texttt{gyro\_z}). The text modality is the ordered fine-grained action sequence inside the coarse segment.
\textbf{Data source:} \url{https://holoassist.github.io/}.

\paragraph{Atomic instance and label.}
An atomic instance is one coarse-grained action segment with duration between 20 and 80 seconds. The model performs binary correctness judgment: \emph{all contained fine-grained actions are correct} (A) versus \emph{at least one contained fine-grained action is incorrect} (B). The final export contains 5{,}936 instances (4{,}991 train / 945 test).

\paragraph{Processing pipeline.}
(1) Load the released annotations and enumerate all coarse-grained action segments.
(2) Keep only coarse segments with duration between 20 and 80 seconds.
(3) Retain segments containing at least one fine-grained action midpoint inside the coarse interval.
(4) Assign the binary label by checking whether any contained fine-grained action has \texttt{Action Correctness} different from \texttt{Correct Action}.
(5) Slice the synchronized eye, head-pose, and gyroscope streams over the same time window, then trim them to a common length for one-to-one temporal alignment.
(6) Build the prompt from the cropped video, ordered fine-grained action sequence, and aligned sensor traces.

\begin{tcolorbox}[title={HoloAssist (concrete instance)},breakable]
\textbf{task\_type:} QA (binary correctness) \quad \textbf{video:} egocentric RGB clip, $454\times 256$, 24.46 fps, 25.55\,s, 625 frames. \quad \textbf{ts\_shape:} $755\times 9$. \\[0.4em]
\begin{center}
  \includegraphics[width=0.8\linewidth]{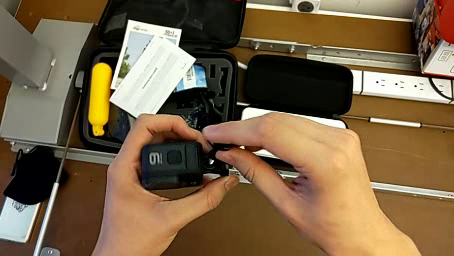}
\end{center}
\textbf{coarse action:}
\begin{tcolorbox}[colback=white, colframe=black!20, boxrule=0.5pt, arc=2pt]
\footnotesize
Action sentence: \emph{The student assembles the mounting\_peg.} \\
Verb: \texttt{assemble}. \quad Adjective: \texttt{none}. \quad Noun: \texttt{mounting\_peg}.
\end{tcolorbox}
\textbf{fine-grained action sequence (10 ordered actions):}
\begin{tcolorbox}[colback=white, colframe=black!20, boxrule=0.5pt, arc=2pt]
\footnotesize
1. rotate gopro \quad 2. open leg \quad 3. approach mounting\_peg \quad 4. grab mounting\_peg \quad 5. insert mounting\_peg \quad 6. grab screw \quad 7. insert screw \quad 8. screw screw \quad 9. rotate gopro \quad 10. slide mounting\_peg.
\end{tcolorbox}
\textbf{variable definitions:}
\begin{tcolorbox}[colback=white, colframe=black!20, boxrule=0.5pt, arc=2pt]
\footnotesize
\texttt{gaze\_dir\_\{x,y,z\}}: components of the performer's gaze direction vector. \quad \texttt{tx, ty, tz}: head translation along $x/y/z$ in space. \quad \texttt{gyro\_\{x,y,z\}}: gyroscope angular velocity around each axis (headset rotational motion).
\end{tcolorbox}
\textbf{time series (per-frame, format \texttt{t(s): gaze\_dir\_x=, gaze\_dir\_y=, gaze\_dir\_z=, tx=, ty=, tz=, gyro\_x=, gyro\_y=, gyro\_z=}):}
\begin{tcolorbox}[colback=white, colframe=black!20, boxrule=0.5pt, arc=2pt]
\footnotesize\raggedright
326.600: gaze\_dir\_x=0.490672, gaze\_dir\_y=0.326931, gaze\_dir\_z=$-$0.807686, tx=0.156120, ty=2.915354, tz=$-$0.046323, gyro\_x=$-$0.009175, gyro\_y=$-$0.012515, gyro\_z=0.056830 \\
326.633: gaze\_dir\_x=0.490563, gaze\_dir\_y=0.327070, gaze\_dir\_z=$-$0.807696, tx=0.156341, ty=2.915125, tz=$-$0.046315, gyro\_x=$-$0.032264, gyro\_y=$-$0.004942, gyro\_z=0.033517 \\
\dots \\
351.733: gaze\_dir\_x=0.401040, gaze\_dir\_y=0.370614, gaze\_dir\_z=$-$0.837743, tx=0.321765, ty=2.821166, tz=$-$0.070192, gyro\_x=0.067685, gyro\_y=0.025116, gyro\_z=$-$0.006590
\end{tcolorbox}
\textbf{prompt (abridged):}
\begin{lstlisting}[basicstyle=\ttfamily\footnotesize,breaklines=true,escapeinside={(*}{*)},breakindent=0pt,columns=fullflexible,breakatwhitespace=true]
You are given a question-answering (QA) task for egocentric human
activity understanding. You are provided with a temporally cropped video
segment from the HoloAssist dataset. The video segment corresponds to
one coarse-grained action performed by the wearer. This coarse-grained
action may contain multiple fine-grained actions. You are also provided
with synchronized sensor data aligned to this same coarse-grained action
segment. Your task is to judge whether all contained fine-grained
actions are correct, or whether at least one contained fine-grained
action is incorrect. Do NOT output any analysis. Return only the option
letter.

<video>
This video clip contains only the target coarse-grained action segment.
</video>

<coarse_action> (*\textbf{coarse action}*) </coarse_action>
<fine_actions>  (*\textbf{fine-grained sequence}*) </fine_actions>
<variables>     (*\textbf{variable defs}*) </variables>
<timeseries>    (*\textbf{time series}*)   </timeseries>
\end{lstlisting}
\textbf{question:} Based on the video content, the fine-grained action sequence, and the sensor data above, does this coarse-grained action segment contain any incorrect fine-grained action? \\
\textbf{options:} A. All contained fine-grained actions are correct. \quad B. At least one contained fine-grained action is incorrect. \\
\textbf{expected answer:} A.
\end{tcolorbox}

\subsection{Healthcare: MIMIC}

\paragraph{Source and modalities.}
The MIMIC collection is a suite of large-scale, de-identified critical-care datasets released via PhysioNet~\citep{pollard_physionet_2026}, spanning complementary clinical modalities collected in routine hospital practice. \textbf{MIMIC-IV}~\citep{Johnson2023MIMIC,PhysioNet-mimiciv-3.1} provides longitudinal, patient- and encounter-level structured EHR data from ICU and hospital stays, including demographics, admissions and transfers, diagnoses and procedures, medication prescriptions and administrations, laboratory measurements, and detailed time-stamped event records. \textbf{MIMIC-IV-Note}~\citep{PhysioNet-mimic-iv-note-2.2} releases free-text clinical narratives, such as discharge summaries and other note types, aligned to hospital encounters. \textbf{MIMIC-CXR-JPG}~\citep{johnson2019mimic,PhysioNet-mimic-cxr-jpg-2.1.0} provides chest radiographs in JPG format together with metadata, enabling joint study of imaging content and patient episodes when used with the structured and textual components.
\textbf{Data sources:} MIMIC-IV (\url{https://physionet.org/content/mimiciv/3.1/}), MIMIC-IV-Note (\url{https://physionet.org/content/mimic-iv-note/2.2/}), MIMIC-CXR-JPG (\url{https://physionet.org/content/mimic-cxr-jpg/2.1.0/}).

\paragraph{Atomic instance and label.}
An atomic instance is anchored at one of two clinically meaningful time points: an imaging-study event (e.g., a chest X-ray acquisition) or a fixed 48-hour mark after hospital admission. The visual stream is the chest X-ray at the anchor; the temporal stream is six routinely recorded vital signs (heart rate, respiratory rate, SpO$_2$, temperature, NIBP systolic, NIBP diastolic) over the observation window; the text modality is the radiology report aligned to the same imaging study. Two prognosis-style binary tasks are evaluated: \emph{in-hospital mortality prediction} (using the first 48 hours after admission, predict whether the patient ultimately dies during the hospital stay) and \emph{discharge readiness prediction} (using a given 48-hour window, predict whether the patient should be discharged within the next 48 hours). The mortality task contains 20 mortality and 130 survival cases (150 samples); the discharge task contains 50 discharge and 100 non-discharge cases (150 samples).

\paragraph{Processing pipeline.}
For each anchor, a fixed observation window of vital-sign time series is extracted prior to the anchor time to provide structured physiologic context. The chest X-ray and the corresponding radiology report associated with the anchor event supply complementary visual and textual evidence aligned to the same encounter. Both tasks are formulated as QA-style multiple-choice prompts requiring integration of physiological measurements with radiology and text documentation.

\begin{tcolorbox}[title={Mimic},breakable]
\textbf{task\_type:} QA \\[0.4em]
\textbf{image:}
\begin{center}
  \includegraphics[width=0.3\linewidth]{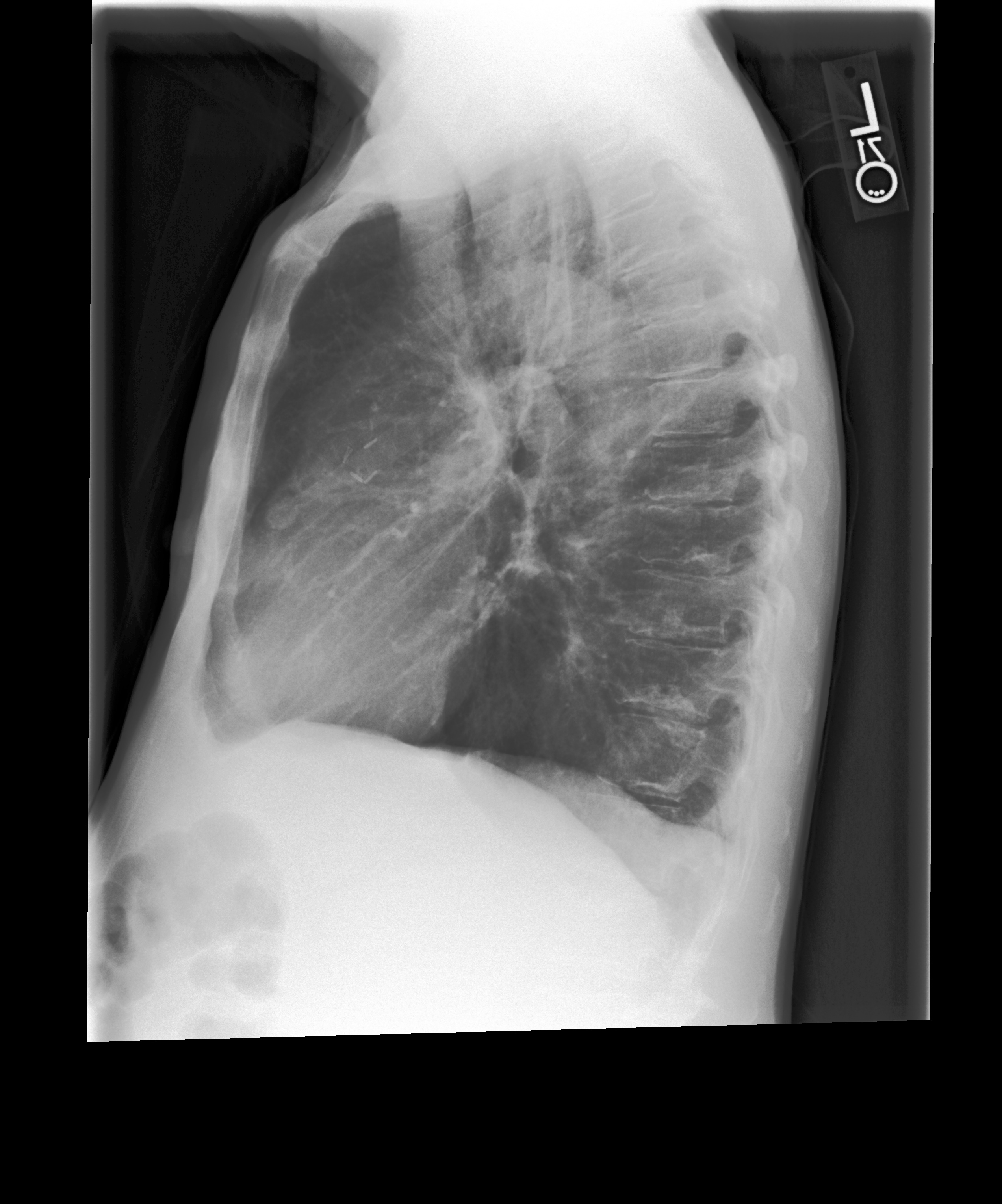}
\end{center}
\textbf{text:}
\begin{tcolorbox}[colback=white, colframe=black!20, boxrule=0.5pt, arc=2pt]
\footnotesize
Radiology report(s) (RR notes) associated with the imaging evidence: \\
{}[RR Note 1 \textbar{} charttime=xxx]: \\
{}[RR Note 2 \textbar{} charttime=xxx]: \\
......
\end{tcolorbox}
\textbf{time series:}
\begin{tcolorbox}[colback=white, colframe=black!20, boxrule=0.5pt, arc=2pt]
\footnotesize
timestamp: \\
heart\_rate\_bpm=, resp\_rate\_insp\_per\_min=, spo2\_percent=, temp\_c=, nbp\_systolic\_mmhg=, nbp\_diastolic\_mmhg=
\end{tcolorbox}
\textbf{prompt\_death:}
\begin{lstlisting}[basicstyle=\ttfamily\footnotesize,breaklines=true,escapeinside={(*}{*)},breakindent=0pt,columns=fullflexible,breakatwhitespace=true]
You are given a question-answering (QA) task. Use the chest X-ray
image(s), radiology report(s), and vital signs as evidence to answer
the multiple-choice question. Do NOT output any analysis. Return only
the option letter.

<image>
MIMIC-CXR-JPG images: each image is a JPG converted from an original
DICOM chest X-ray.
(*\textbf{image}*)
</image>

<report>
Radiology report(s) (RR notes):
(*\textbf{text}*)
</report>

<variables>
Vital sign variables: an introduction to each variable.
</variables>

<timeseries>
Aligned time series (each line is one timestamp; missing values empty):
(*\textbf{time series}*)
</timeseries>
\end{lstlisting}
\textbf{question\_death:} Based on the chest X-ray image(s), radiology report(s), and the ICU first-48-hour vital signs, will the patient die during this hospital admission? \\
\textbf{question\_disch:} Using the chest X-ray image(s), radiology report(s), and the vital signs observed in the 48-hour lookback window ending at the anchor time, will the patient be discharged within the next 48 hours after the anchor time? \\
\textbf{options:} A. Yes \quad B. No
\end{tcolorbox}

\subsection{Recommendation: PixelRec}

\paragraph{Source and modalities.}
PixelRec~\citep{Cheng2023PixelRec} is a real-world recommendation dataset that exposes raw item-side cover images as first-class signals for modeling. Alongside covers, it ships standard recommender-system records, including user--item interaction logs and item metadata (titles, tags, descriptions when available), which together support studying how visual presentation and textual context correlate with user feedback. Unlike benchmarks that only release pre-extracted visual embeddings, PixelRec makes the original cover imagery directly accessible.
\textbf{Data source:} \url{https://github.com/westlake-repl/PixelRec}.

\paragraph{Atomic instance and task.}
An atomic instance is one item. Item-level forecasting instances are constructed by aggregating raw user--item interactions into a daily time series per item (daily interaction counts). To reduce irregularity from sparsity, a long consecutive segment of daily observations is retained per item, overly long histories are truncated to a fixed maximum length, and the last several days are reserved as forecast targets. Each instance combines three modalities: time series (historical daily counts as the primary signal), text (concatenation of title, tag, and description), and a single cover image. The model is asked to forecast the next 6 days of interaction counts.

\paragraph{Processing pipeline.}
A forecasting-oriented prompt is used that conditions on (i) item metadata, (ii) the cover image, (iii) the historical daily interaction series as ordered \texttt{(timestamp, value)} pairs, and (iv) a list of future timestamps to be predicted.

\begin{tcolorbox}[title={PixelRec},breakable]
\textbf{task\_type:} prediction \\[0.4em]
\textbf{image:}
\begin{center}
  \includegraphics[width=0.4\linewidth]{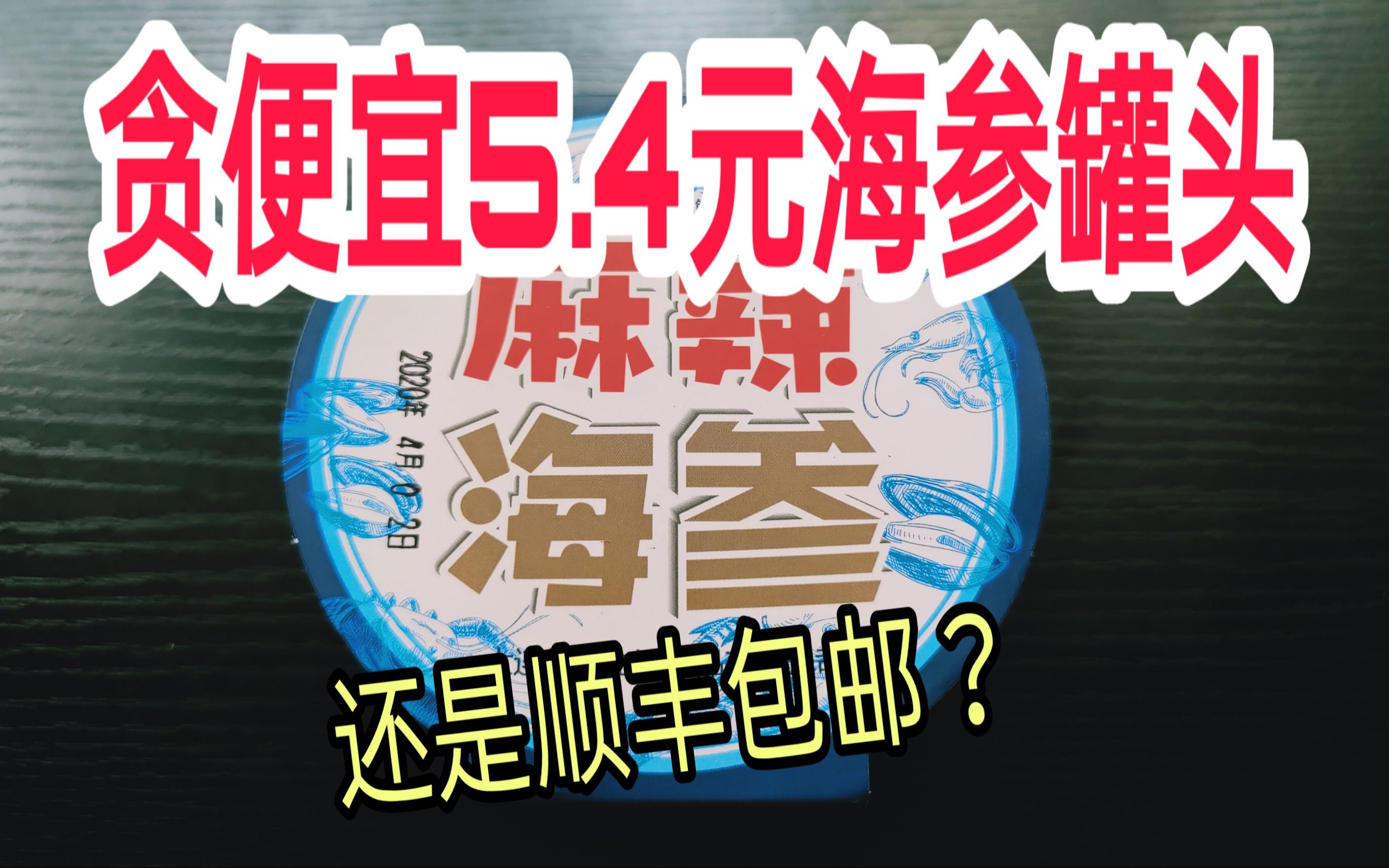}
\end{center}
\textbf{text:}
\begin{tcolorbox}[colback=white, colframe=black!20, boxrule=0.5pt, arc=2pt]
\footnotesize
title: xxx \quad tag: xxx \quad description: xxx
\end{tcolorbox}
\textbf{time series (time, interaction volume):}
\begin{tcolorbox}[colback=white, colframe=black!20, boxrule=0.5pt, arc=2pt]
\footnotesize
(2021-11-24, 41) \\
(2021-11-25, 26) \\
(2021-11-26, 11) \\
(2021-11-27, 5) \\
(2021-11-28, 5) \\
......
\end{tcolorbox}
\textbf{prompt:}
\begin{lstlisting}[basicstyle=\ttfamily\footnotesize,breaklines=true,escapeinside={(*}{*)},breakindent=0pt,columns=fullflexible,breakatwhitespace=true]
I have a time series forecasting task for you.
Here is some context about the item (title/tag/description).
<context>
(*\textbf{text}*)
</context>

Here is the cover of the item.
(*\textbf{image}*)

Here is a historical time series. Timestamps are daily,
values are daily interaction counts.
<history>
(*\textbf{time series}*)
</history>
Now please predict the value at the following timestamps:
Return the forecast in (timestamp, value) format in between
<forecast> and </forecast> tags.
Do not include any other information (e.g., comments) in the forecast.
\end{lstlisting}
\textbf{task:} Predict the next 6 time points of the interaction volume.
\end{tcolorbox}

\subsection{Finance: FinMultiTime}

\paragraph{Source and modalities.}
FinMultiTime~\citep{Xu2025FinMultiTime} is a multimodal financial forecasting dataset centered on market time series as prediction targets (e.g., stock price/return movements). It pairs the core numerical signals with heterogeneous auxiliary evidence, including textual financial news and visual K-line/candlestick charts, enabling research on how narrative events and visualized technical patterns complement numerical dynamics. The corpus covers equities from the China A-share market and the S\&P 500 universe, supporting cross-market evaluation under diverse regimes.
\textbf{Data source:} \url{https://huggingface.co/datasets/Wenyan0110/Multimodal-Dataset-Image_Text_Table_TimeSeries-for-Financial-Time-Series-Forecasting}.

\paragraph{Atomic instance and task.}
The MMTA subset is built on the S\&P 500 universe. An atomic instance is one asset paired with a half-year history window. Given a fixed history interval (e.g., the first half of a year), the model predicts the closing prices for the subsequent horizon. Each instance includes three modalities: the historical close-price sequence within the half-year window (numeric), a candlestick chart summarizing the same period (visual), and a small set of recent news snippets selected strictly from before the prediction timestamps (text).

\paragraph{Processing pipeline.}
Forecasting-style prompts are used; they condition on the aligned multimodal evidence (price history, candlestick image, and pre-horizon news) and ask the model to predict closing prices at specified future dates. The model outputs 12 trading-day forecasts per instance.

\begin{tcolorbox}[title={FinMultiTime},breakable]
\textbf{task\_type:} prediction \\[0.4em]
\textbf{image:}
\begin{center}
  \includegraphics[width=0.5\linewidth]{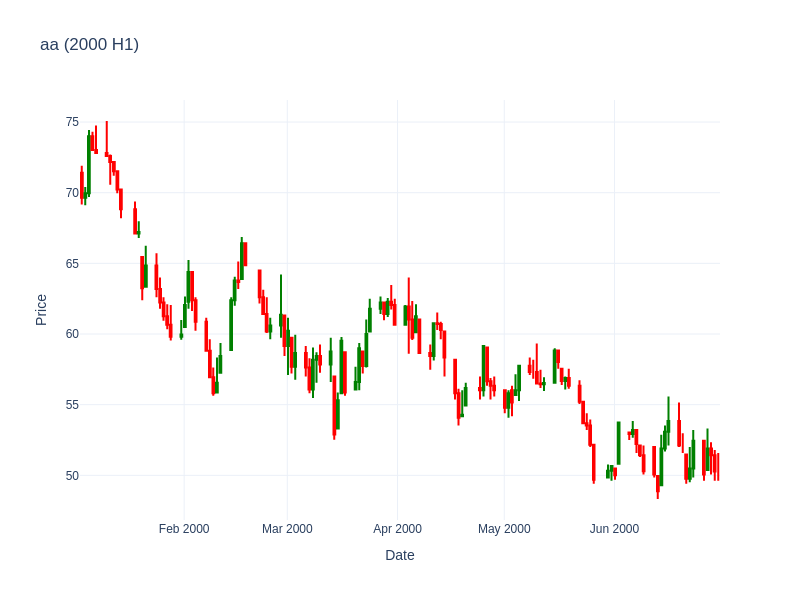}
\end{center}
\textbf{text:}
\begin{tcolorbox}[colback=white, colframe=black!20, boxrule=0.5pt, arc=2pt]
\footnotesize
Recent news summaries (most recent last): \\
......
\end{tcolorbox}
\textbf{time series:}
\begin{tcolorbox}[colback=white, colframe=black!20, boxrule=0.5pt, arc=2pt]
\footnotesize
(2024-01-02, 168.8) \\
(2024-01-03, 165.2) \\
(2024-01-04, 165.1) \\
......
\end{tcolorbox}
\textbf{prompt:}
\begin{lstlisting}[basicstyle=\ttfamily\footnotesize,breaklines=true,escapeinside={(*}{*)},breakindent=0pt,columns=fullflexible,breakatwhitespace=true]
I have a time series forecasting task for you.
Here is some context and a candlestick chart image about the task.
Make sure to factor in any background knowledge, satisfy any
constraints, and respect any scenarios.
<context>
(*\textbf{text}*)
</context>

<image>
Candlestick image: this image shows the historical candlestick chart
for the asset.
(*\textbf{image}*)
</image>

<history>
Here is a historical time series in (timestamp, value) format:
(*\textbf{time series}*)
</history>

Now please predict the value at the following timestamps: xxx
Return the forecast in (timestamp, value) format in between
<forecast> and </forecast> tags.
Do not include any other information (e.g., comments) in the forecast.
\end{lstlisting}
\textbf{task:} Predict the next 12 time points of the closing price.
\end{tcolorbox}

\subsection{Climate: Terra}

\paragraph{Source and modalities.}
Terra~\citep{Chen2024Terra} is a global-scale multimodal spatio-temporal dataset constructed on a 1$^{\circ}$ latitude--longitude grid (64{,}800 spatial cells). It provides long-horizon climate time series from 1979-01-01 to 2024-01-01 at three temporal resolutions (Monthly, Daily, 3-Hourly). Time-series variables include precipitation, temperature, surface pressure, relative humidity, specific humidity, wind speed, and downward shortwave and longwave radiation. Beyond numerical climate fields, Terra provides aligned image modalities for each spatial cell, including satellite imagery and geophysical map layers such as geoid, free-air anomaly (FAA), magnetic anomaly (MA), land--sea mask, relief, and vertical gravity gradient (VGG). It also exposes two levels of text: (i) per-cell metadata (coordinates, city, vegetation, climate, elevation), and (ii) LLM-generated descriptions that expand the metadata into richer semantic narratives.
\textbf{Data source:} \url{https://github.com/CityMind-Lab/NeurIPS24-Terra?tab=readme-ov-file}. \quad
\textbf{Paper:} \url{https://openreview.net/pdf?id=I0zpivK0A0}.

\paragraph{Atomic instance and task.}
An atomic instance is one grid cell. At each location, a daily time series is extracted and a 120-day window is formed; the first 100 days are used as historical context and the next 20 days are forecast. To construct a tractable subset, valid grid locations are randomly sampled over the globe under basic geographic constraints on latitude and longitude. Each sampled location yields one multimodal instance, which encourages models to learn patterns that generalize across regions. The remote-sensing imagery and related geospatial raster products provide static or slowly varying spatial context. The location-conditioned LLM-generated descriptions provide a compact semantic prior that complements numeric history and imagery.

\paragraph{Processing pipeline.}
Prediction-style prompts are used: given the aligned multimodal context (historical daily series, associated imagery, location text), the model outputs the next 20 daily values for all variables in the same order.

\begin{tcolorbox}[title={Terra},breakable]
\textbf{task\_type:} prediction \\[0.4em]
\textbf{image:}
\begin{center}
  \includegraphics[width=0.5\linewidth]{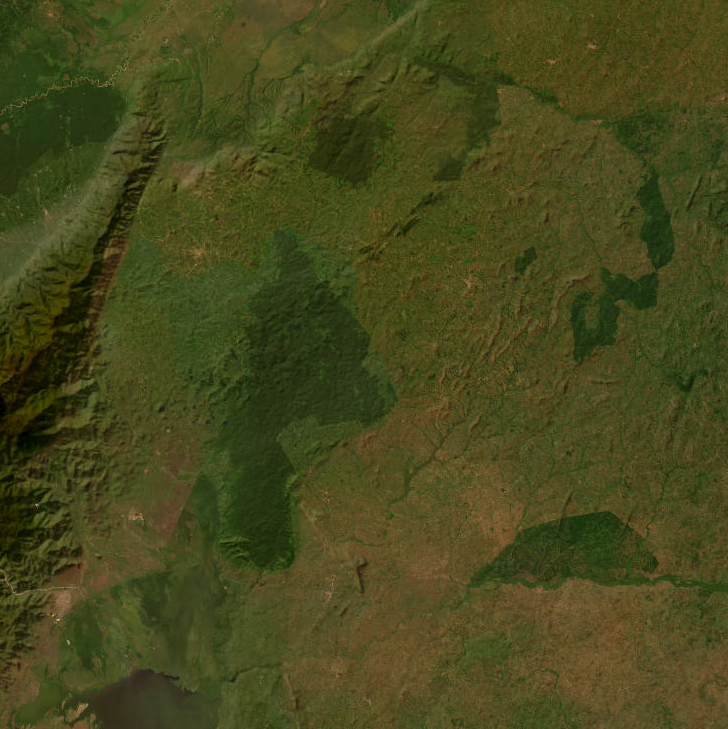}
\end{center}
\textbf{text:}
\begin{tcolorbox}[colback=white, colframe=black!20, boxrule=0.5pt, arc=2pt]
\footnotesize
LLM-generated description of the grid cell, expanded from the dataset's
per-cell metadata (coordinates, city, vegetation, climate, elevation). \\
......
\end{tcolorbox}
\textbf{time series (example):}
\begin{tcolorbox}[colback=white, colframe=black!20, boxrule=0.5pt, arc=2pt]
\footnotesize
timestamp: lwd, precipitation, pres, relhumidity, spechumidity, temp, wind \\
......
\end{tcolorbox}
\textbf{prompt:}
\begin{lstlisting}[basicstyle=\ttfamily\footnotesize,breaklines=true,escapeinside={(*}{*)},breakindent=0pt,columns=fullflexible,breakatwhitespace=true]
You are given a geospatial multivariate time series forecasting task on
the Terra 1*1 global grid. We focus on a single grid cell identified by:
- Grid label: xxx

<image>
Each image is aligned to the same grid cell "{label}" and provides
complementary static geophysical context.
(*\textbf{image}*)
</image>

You are given multiple historical variables for the same grid cell and
timestamps.
(*\textbf{A description of each variable}*)
All variables are aligned on the same timestamps and must be forecast
in the SAME order.

<text>
The following description is generated from metadata and provides
semantic/geographic context for the same grid cell:
(*\textbf{text}*)
</text>

Given the historical multivariate time series, predict the future
values at the next timestamps xxx. You must output forecasts for ALL
variables listed below, in the SAME order.

Variable order:
xxx

Historical time series:
<history>
(*\textbf{time series}*)
</history>

Forecast timestamps:
xxx

## Output format
Return ONLY the forecast lines between <forecast> and </forecast> tags.
Each forecast line must be:
(timestamp: value_1, value_2, value_3, ...)
- Use the SAME variable order as specified above.
- Output exactly xxx lines.
- Do NOT include any extra text, explanations, or comments.
\end{lstlisting}
\textbf{task:} Predict the next 20 time points of all variables.
\end{tcolorbox}

\subsection{Dataset Licenses and Terms of Use}
\label{app:dataset-licenses}

Table~\ref{tab:dataset_licenses} lists, for each of the nine MMTA source datasets, the upstream license or access terms, the access URL used in this work, the use restriction that applies to MMTA's reuse, and how the domain is distributed in the public release. The release contains test splits only. It hosts the 1{,}322 test records of AgiBotWorld, Future Factories, HoloAssist, and Terra, together with 1{,}400 alternative Terra channel views that reuse the same 200 instances and images. For CoVLA, CUHK-X, MIMIC, PixelRec, and FinMultiTime, the release contains processing code only: users obtain the source data under the upstream terms (for MIMIC, through their own credentialed PhysioNet account) and build the records locally, and neither source data nor generated records are redistributed. Locally rebuilt records can differ from the benchmark selection when the available upstream versions differ. Original MMTA contributions are released under CC BY 4.0, which covers only rights held by the MMTA authors. Hosted source material keeps its upstream license, and the license texts ship with the release, so the AgiBotWorld and Future Factories subsets stay under share-alike terms and the AgiBotWorld and Terra subsets stay non-commercial. Third-party processing code keeps its existing notices, and its provenance is documented in the release. The TimeOmni-v code is released under Apache-2.0 in its public repository.

\begin{table*}[t]
\centering
\footnotesize
\setlength{\tabcolsep}{4pt}
\renewcommand{\arraystretch}{1.15}
\caption{Upstream license / access terms for the nine MMTA source datasets and their distribution in the public MMTA release. The ``Use restriction'' column summarizes the most stringent term that applies to MMTA's reuse; readers should consult the original license text before any further redistribution or commercial use. ``Hosted'' gives the number of test records in the release; ``Code only'' means that the release contains processing code but no source data or generated records.}
\label{tab:dataset_licenses}
\begin{tabular}{@{}>{\raggedright\arraybackslash}p{0.15\linewidth} >{\raggedright\arraybackslash}p{0.21\linewidth} >{\raggedright\arraybackslash}p{0.17\linewidth} >{\raggedright\arraybackslash}p{0.25\linewidth} >{\raggedright\arraybackslash}p{0.13\linewidth}@{}}
\toprule
\textbf{Dataset} & \textbf{Source} & \textbf{License / access} & \textbf{Use restriction} & \textbf{MMTA release} \\
\midrule
AgiBotWorld~\citep{AgiBotWorld2026} & Hugging Face: \texttt{agibot-world/\allowbreak AgiBotWorld2026} (tasks 3641, 3777) & CC BY-NC-SA 4.0 & Non-commercial only; attribution and share-alike required for adaptations. & Hosted (131) \\
CoVLA~\citep{CoVLA2024} & Hugging Face (gated): \texttt{turing-motors/\allowbreak CoVLA-Dataset} & Custom CoVLA terms; text descriptions also under VideoLLaMA~2 license & Academic / non-commercial only; non-transferable, non-redistributable; confidentiality required. & Code only \\
CUHK-X~\citep{CUHKX2024} & GitHub: \texttt{openaiotlab/CUHK-X} & Code MIT; data under DUA (non-commercial research, no redistribution); derived annotations CC BY 4.0 & Raw data is \emph{not} MIT; follow the DUA and do not redistribute the source data. & Code only \\
Future Factories~\citep{Harik2024FutureFactories} & Kaggle: \texttt{ramyharik/\allowbreak ff-2024-08-13-\allowbreak multi-modal-\allowbreak dataset-33} (version 1) & CC BY-SA 4.0 & Attribution and share-alike required for adaptations. & Hosted (46) \\
HoloAssist~\citep{HoloAssist2023} & \url{https://holoassist.github.io/} & CDLA-Permissive 2.0 & Permissive data license; the agreement text ships with shared data and modifications. & Hosted (945) \\
MIMIC (IV / CXR-JPG / IV-Note)~\citep{Johnson2023MIMIC,PhysioNet-mimiciv-3.1,johnson2019mimic,PhysioNet-mimic-cxr-jpg-2.1.0,PhysioNet-mimic-iv-note-2.2} & PhysioNet & PhysioNet Credentialed Health Data License 1.5.0 & Credentialed access only; no re-identification; access not transferable; HIPAA training and code-sharing obligations apply. & Code only \\
PixelRec~\citep{Cheng2023PixelRec} & GitHub: \texttt{westlake-repl/\allowbreak PixelRec} & Code MIT; data under custom non-commercial research and education license & Data is \emph{not} MIT; no commercial use; no rights to copy, modify, publish, or redistribute the dataset itself. & Code only \\
FinMultiTime~\citep{Xu2025FinMultiTime} & Hugging Face: \texttt{Wenyan0110/\allowbreak Multimodal-\allowbreak Dataset-...} & MIT (per dataset metadata) & Preserve copyright and license notice on substantial redistribution. & Code only \\
Terra~\citep{Chen2024Terra} & Hugging Face: \texttt{onedean/Terra}; project: \texttt{CityMind-Lab/\allowbreak NeurIPS24-Terra} & CC BY-NC 4.0 & Non-commercial only. Per the Terra paper, the satellite tiles come from ArcGIS (Esri), whose terms permit non-commercial redistribution including research use, and the map layers are rendered with PyGMT from GMT remote datasets. & Hosted (200; 1{,}400 channel views) \\
\bottomrule
\end{tabular}
\end{table*}

\section{Modality Coverage and Data Quality}
\label{app:data-quality}

This appendix supports Section~\ref{sec:bench-quality}. It states which domains carry sample-level text, how the labels are derived, how the two generated components were audited, how video and time series were checked for alignment, and how the domains whose modalities share a source should be read.

\subsection{Sample-Level Text}
\label{app:sample-text}

Text is counted as a sample-level modality only when it varies across samples and carries scene, event, or background information about the sample. Fixed task instructions, field definitions, and label lists define the task and are not counted. A field-level check of the training data shows that seven domains carry such text in every sample (Table~\ref{tab:sample_text}). In CoVLA and CUHK-X the prompt contains only the fixed task description, the variable or sensor definitions, and the label space, so these two domains are video$+$TS tasks with respect to sample-level evidence.

\begin{table}[h]
\centering
\small
\caption{Sample-level text in each MMTA domain.}
\label{tab:sample_text}
\begin{tabular}{@{}lcp{0.62\linewidth}@{}}
\toprule
\textbf{Domain} & \textbf{Sample-level text} & \textbf{Per-sample text content} \\
\midrule
AgiBotWorld      & yes & Upstream task instruction and the time-ordered execution steps inside the Task Frame \\
CoVLA            & no  & Fixed driving-task description, variable definitions, question, and three-way label space \\
CUHK-X           & no  & Fixed HAR task description, sensor definitions, and 40-way action space \\
Future Factories & yes & Production-cycle report (cycle ID, UTC time range, record count, process, signal meaning, camera notes) \\
HoloAssist       & yes & Coarse-grained action description and the time-ordered fine-grained actions in the segment \\
MIMIC            & yes & Radiology report of the imaging study (findings, and possibly clinical history and prior-exam comparisons) \\
PixelRec         & yes & Item title, tags, and description \\
FinMultiTime     & yes & News text about the asset \\
Terra            & yes & Geographic and climatic description generated from the grid-cell metadata \\
\bottomrule
\end{tabular}
\end{table}

\subsection{Label Derivation}
\label{app:label-derivation}

Only the CoVLA brake-reason label is generated by our pipeline; the other three relabeled domains keep the upstream label semantics through deterministic, traceable steps.

\paragraph{AgiBotWorld.} The upstream \texttt{Error Frame} annotation provides an \texttt{error\_cause}. Error Frames whose temporal midpoint falls inside a Task Frame are collected, and the raw descriptions are normalized to \emph{paused}, \emph{meaningless operation}, \emph{fall}, or \emph{other}. Task Frames with more than one normalized anomaly type are excluded rather than collapsed into one label.

\paragraph{Future Factories.} The upstream cycle-management records give the missing-part condition that was set by hand in each experiment. Records are joined to production cycles by cycle ID and UTC time range, and the upstream descriptions are normalized to the four labels. The label is therefore a mapping of the experimental condition, not an inference from the input time series.

\paragraph{HoloAssist.} The upstream annotation gives an \texttt{Action Correctness} value for every fine-grained action. Fine-grained actions are assigned to coarse-grained segments by temporal containment; a segment is labeled A if all its actions are \texttt{Correct Action} and B otherwise.

\paragraph{CoVLA.} Braking intervals are located from the upstream \texttt{brakePressed} signal (at least six consecutive frames). Inside an interval, label A is assigned if \texttt{lead\_a}$<0$ holds for at least two consecutive frames, label B if a left or right turn signal is on for at least two consecutive frames, and label C otherwise; intervals that trigger both A and B are excluded.

\subsection{Human Audit of the Generated Components}
\label{app:human-audit}

The CoVLA rule-derived labels and the Terra descriptions, which the Terra authors generated with an LLM, were audited because they are the two components that need semantic verification. For each, 100 samples were drawn at random and checked independently by two annotators; every disagreement was resolved by adjudication, and the adjudicated decision was used to judge the generated content. The two annotators are doctoral students with at least one year of research experience in time-series processing, forecasting, or multimodal time-series analysis. Table~\ref{tab:human_audit} reports the results. The identified items are scheduled for correction in the next dataset release; no sample was deleted, because the underlying video, images, and time series remain usable and aligned.

\begin{table}[h]
\centering
\small
\setlength{\tabcolsep}{4pt}
\caption{Double-annotated audit of the two generated components. Agreement and Cohen's $\kappa$ are computed before adjudication; accuracy is computed after adjudication, with a 95\% Wilson interval.}
\label{tab:human_audit}
\resizebox{\linewidth}{!}{\begin{tabular}{@{}lcccccc@{}}
\toprule
\textbf{Component} & \textbf{N} & \textbf{Agreement} & \textbf{$\kappa$} & \textbf{Adjudicated} & \textbf{Accuracy (95\% CI)} & \textbf{Items to correct} \\
\midrule
CoVLA brake-reason labels & 100 & 94\% & 0.910 & 6 & 96\% (90.2--98.4\%) & 4 \\
Terra generated text      & 100 & 95\% & 0.687 & 5 & 93\% (86.3--96.7\%) & 7 \\
\bottomrule
\end{tabular}}
\end{table}

\paragraph{Error types.} The CoVLA errors concern fine-grained driving semantics: at road crossings with a lead vehicle present, the rule can attribute the braking to lead-vehicle deceleration, and some lane changes are labeled as turns. The Terra errors concentrate on ocean cells, where the text can contain internal geographic contradictions or read a negative elevation (sea-floor depth) as a land altitude.

\subsection{Cross-Modal Alignment Audit}
\label{app:alignment-audit}

The mechanical audit checks source-sample identity, timestamps or frame boundaries, time-series coverage, and file existence for all 25{,}057 video$+$TS samples. Because clips are cut with keyframe-level stream copy, an exact frame-count match cannot separate a lost target event from a few extra context frames, so two rates are reported. The \emph{coverage failure rate} is the fraction of samples whose released clip does not fully contain the target annotation window. The \emph{boundary overhang rate} is the fraction of released frames that lie outside the target window.

\begin{table}[h]
\centering
\small
\caption{Mechanical alignment audit of the video$+$TS domains.}
\label{tab:alignment_audit}
\begin{tabular}{@{}lcccc@{}}
\toprule
\textbf{Domain} & \textbf{Samples} & \textbf{Coverage failure} & \textbf{Boundary overhang} & \textbf{Max.\ overhang} \\
\midrule
AgiBotWorld      & 1{,}308  & 0 (0.000\%) & 2.35\% & 66 frames / 2.20\,s \\
CoVLA            & 14{,}781 & 0 (0.000\%) & 4.80\% & 11 frames / 0.55\,s \\
CUHK-X           & 2{,}937  & 0 (0.000\%) & 0.00\% & 0 \\
Future Factories & 95       & 0 (0.000\%) & 0.00\% & 0 \\
HoloAssist       & 5{,}936  & 4 (0.067\%) & 0.55\% & 13 frames / ${\approx}$0.5\,s \\
\midrule
Total            & 25{,}057 & 4 (0.016\%) & 1.42\% & --- \\
\bottomrule
\end{tabular}
\end{table}

The four coverage failures are HoloAssist samples whose annotated end lies 3--4 frames past the end of the source video, so no existing content was lost in cutting. The video$+$TS data are therefore aligned at the level of the annotated segment, with a small amount of boundary context set by the video encoding. The image$+$TS domains are joined on deterministic keys: patient, imaging study, and admission time for MIMIC; item ID for PixelRec; asset and date for FinMultiTime; and grid cell and date for Terra.

\subsection{CoVLA Labels and the Input Time Series}
\label{app:covla-shortcut}

The CoVLA label rule and the released time series overlap only partially. The variables that locate the braking interval (\texttt{brakePressed}) and that define class B (the turn signals) are not among the five released channels (\texttt{vEgo}, \texttt{aEgo}, \texttt{steeringAngleDeg}, \texttt{lead\_speed\_kmh}, \texttt{lead\_a}); only \texttt{lead\_a} enters the class-A rule. The rule fires on a local trigger of two consecutive frames, whereas the released series covers the whole braking interval of 6--600 steps. A local deceleration also does not separate road crossings, traffic signals, and lead-vehicle behavior, which the video resolves. Consistent with this, the TS-only variant of TimeOmni-v trails the full model by 16.4\% on CoVLA (Figure~\ref{fig:modality}).

\subsection{Domains with Shared Sources}
\label{app:shared-sources}

Non-redundancy is read here as a property of the information each modality carries, not of how the modalities were acquired. Three domains combine modalities that share a source.

\paragraph{FinMultiTime.} The candlestick chart and the price series come from the same market data. The released series carries only the close price, whereas the chart also encodes the open, high, and low prices and thus shows range, gaps, and volatility; the two share the close price. The news text adds company events that the price series does not contain.

\paragraph{Terra.} The text is generated from the grid-cell metadata (coordinates, country or city, land cover, climate type, elevation) and discusses terrain, ocean influence, monsoons, prevailing winds, vegetation, and human activity. These static descriptors are not encoded in the seven forecast channels, and the images add satellite appearance and geophysical fields. The three modalities are linked through the same grid cell but do not substitute for one another.

\paragraph{MIMIC.} The chest X-ray and the radiology report come from the same imaging study. The image keeps the raw morphology and spatial detail, while the report records the radiologist's interpretation and may add clinical history and comparisons with prior exams, which cannot be read from the pixels. The vital signs of the first 48 hours form a separate physiological stream, and the target is a later hospital outcome. The training-time ablation shows that removing the vital signs does not lower Macro-F1 on MIMIC (Figure~\ref{fig:modality}), so MIMIC is a boundary case of complementarity rather than strong evidence of it.

\section{TimeOmni-v Training Details}
\label{app:TimeOmni-v-training}

This appendix documents the training recipe used to fine-tune \textbf{TimeOmni-v} (Section~\ref{sec:TimeOmni-v}) on each MMTA dataset. A single recipe is shared across the nine MMTA datasets, with only two hyperparameters varying per dataset. All experiments are run from the same base model and the same set of trainable modules; the resulting checkpoints are evaluated under the protocol of Section~\ref{sec:evaluated-models}.

\subsection{Initialization and trainable modules}
\label{app:TimeOmni-v-training-init}

TimeOmni-v is initialized from two pretrained backbones. The Omni-LLM half is initialized from Qwen2.5-Omni-7B~\citep{Qwen25Omni} (Thinker half), and the time-series encoder branch is initialized from Chronos-2~\citep{Chronos2}. The vision and audio towers of the Thinker are kept frozen, since the visual evidence in MMTA is already covered by the pretrained vision encoder and the audio modality is not used. Four sets of parameters are fine-tuned in full precision: (i) the Chronos-2 encoder; (ii) the lightweight TS-to-LLM adapter that maps Chronos-2 patch embeddings to the LLM hidden size; (iii) the embedding rows of the three new TS special tokens (\texttt{<|ts\_start|>}, \texttt{<|ts\_placeholder|>}, \texttt{<|ts\_end|>}); and (iv) the forecast head, attached only on forecasting datasets. The Omni-LLM itself is fine-tuned with LoRA~\citep{Hu2022LoRA} of rank 8, $\alpha=32$, and dropout $0.05$, applied to all attention projections (\texttt{q\_proj}, \texttt{k\_proj}, \texttt{v\_proj}, \texttt{o\_proj}) and dense MLP projections (\texttt{gate\_proj}, \texttt{up\_proj}, \texttt{down\_proj}) of every transformer block.

For each MMTA dataset, an independent TimeOmni-v adapter set is trained on top of the same base model. An adapter set comprises the LoRA weights, the Chronos-2 encoder, the TS-to-LLM adapter, the new TS-token embedding rows, and a forecast head where applicable.

\subsection{Optimization}
\label{app:TimeOmni-v-training-opt}

The model is optimized with AdamW ($\beta_1=0.9$, $\beta_2=0.999$, $\varepsilon=10^{-8}$, weight decay $0$). A linear warmup is applied for the first $10\%$ of the training steps, followed by a cosine learning-rate decay schedule. Gradients are clipped at max-norm $1.0$. Training is run in bf16 with gradient checkpointing enabled, on top of DeepSpeed ZeRO stage 2.

To keep per-step token counts balanced across DDP ranks under variable sequence length, batches are formed by a token-budget dynamic sampler that greedily packs each rank's batch up to ${\sim}40\,000$ padded tokens, with a hard cap of $32$ samples per rank-batch. The gradient-accumulation is $1$.  All experiments are run on NVIDIA H200 140\,GB GPUs, with the GPU count per run scaled to the size of the dataset (1--8 GPUs per dataset).

\subsection{Per-dataset hyperparameters}
\label{app:TimeOmni-v-training-hparams}

Only the learning rate and the number of training epochs vary across MMTA datasets; all other hyperparameters follow Sections~\ref{app:TimeOmni-v-training-init} and \ref{app:TimeOmni-v-training-opt}. Table~\ref{tab:TimeOmni-v_train_hparams} summarizes the per-dataset settings.

\paragraph{Learning rate.} A single rule is used to pick the per-dataset learning rate. The learning rate is initialized at $1\times10^{-5}$. A short pilot run is then launched, and the training loss is monitored. If the loss diverges to NaN within the first epoch, the learning rate is halved and the pilot is restarted; this halve-and-retry is repeated until the loss stays finite throughout the pilot. The smallest stable rate found this way is then used for the full training run. Forecasting variants tend to settle at a smaller rate than classification ones, since the freshly initialized forecast head amplifies the early gradient signal flowing back into the LoRA-side parameters.

\paragraph{Number of epochs.} The epoch count is set to keep each dataset's optimizer-step budget comparable. Concretely, the number of epochs is chosen so that the total number of optimizer steps per dataset stays above $400$, and is otherwise increased only as needed to compensate for very small training sets. As a result, datasets with large training sets (\texttt{covla}, \texttt{holo\_assist}) use short schedules of $10$ epochs, while datasets with small training sets (\texttt{future\_factories}, \texttt{mimic\_death}, \texttt{mimic\_disch}, \texttt{pixelrec\_perch}) extend to $30$--$100$ epochs to clear the same step floor.

\begin{table}[t]
\centering
\small
\setlength{\tabcolsep}{6pt}
\caption{Per-dataset TimeOmni-v training hyperparameters. All other settings (initialization, frozen/trainable modules, optimizer, schedule shape, batching, hardware) are shared across datasets and described in Sections~\ref{app:TimeOmni-v-training-init} and \ref{app:TimeOmni-v-training-opt}.}
\label{tab:TimeOmni-v_train_hparams}
\begin{tabular}{@{}llllc@{}}
\toprule
\textbf{Dataset} & \textbf{Modalities} & \textbf{Task} & \textbf{Learning rate} & \textbf{Epochs} \\
\midrule
\texttt{covla}             & video $+$ TS & classification & $1\times 10^{-5}$ & 10 \\
\texttt{agibot}            & video $+$ TS & classification & $1\times 10^{-5}$ & 20 \\
\texttt{cuhk\_x\_har}      & video $+$ TS & classification & $1\times 10^{-5}$ & 10 \\
\texttt{holo\_assist}      & video $+$ TS & classification & $2\times 10^{-6}$ & 10 \\
\texttt{future\_factories} & video $+$ TS & classification & $2\times 10^{-6}$ & 100 \\
\texttt{mimic\_death}      & image $+$ TS & classification & $2\times 10^{-6}$ & 30 \\
\texttt{mimic\_disch}      & image $+$ TS & classification & $2\times 10^{-6}$ & 30 \\
\texttt{terra\_perch}      & image $+$ TS & forecasting    & $5\times 10^{-6}$ & 5 \\
\texttt{sp500\_perch}      & image $+$ TS & forecasting    & $1\times 10^{-5}$ & 5 \\
\texttt{pixelrec\_perch}   & image $+$ TS & forecasting    & $2\times 10^{-6}$ & 30 \\
\bottomrule
\end{tabular}
\end{table}

\section{Additional Results}
\label{app:additional-results}

\subsection{Decomposed Metrics and Failure Taxonomy}
\label{app:decomposed}

Tables~\ref{tab:decomp_cls}--\ref{tab:failure_taxonomy} decompose every cell of Table~\ref{tab:main_results} into the base metric on the parsed subset, the success rate (SR), and the causes of failure. Context overflow appears as CUDA OOM, because an over-length prompt aborts in the first forward pass; semantic errors are the complement of the base metric on the parsed subset. The two MIMIC subtasks are reported separately.

\begin{table}[h]
\centering
\caption{Classification: base Macro-F1 on the parsed subset, with SR (\%) in parentheses. ``---'': no sample was parsable. Best base Macro-F1 per column in \textbf{bold}.}
\label{tab:decomp_cls}
\setlength{\tabcolsep}{3pt}
\resizebox{\linewidth}{!}{\begin{tabular}{l ccccccc}
\toprule
Model & AgiBot (131) & CoVLA (1{,}344) & CUHK-X (308) & FuFa (46) & HoloA (945) & MIMIC-death (75) & MIMIC-disch (75) \\
\midrule
InternVL3.5-4B           & 0.070 (28.2) & 0.495 (100.0) & 0.218 (100.0) & 0.049 (100.0) & 0.514 (49.3) & 0.547 (100.0) & 0.508 (100.0) \\
InternVL3.5-8B           & 0.138 (19.1) & 0.515 (99.8)  & 0.260 (94.5)  & 0.050 (97.8)  & 0.483 (43.7) & 0.558 (100.0) & 0.571 (100.0) \\
Qwen3-VL-8B-Instruct     & 0.139 (77.9) & 0.592 (100.0) & 0.084 (100.0) & 0.049 (100.0) & 0.273 (94.7) & \textbf{0.769} (100.0) & 0.542 (100.0) \\
Qwen3-VL-30B-A3B-Inst.   & --- (0.0)    & 0.577 (100.0) & 0.086 (99.4)  & 0.129 (100.0) & --- (0.0)    & 0.638 (100.0) & 0.532 (100.0) \\
Qwen2.5-Omni-7B          & 0.108 (56.5) & 0.424 (100.0) & 0.099 (100.0) & 0.198 (100.0) & 0.303 (74.6) & 0.559 (100.0) & 0.551 (100.0) \\
Qwen3-Omni-30B-A3B-Inst. & 0.125 (28.2) & 0.491 (100.0) & 0.244 (100.0) & 0.129 (100.0) & 0.331 (63.5) & 0.741 (100.0) & 0.544 (100.0) \\
GPT-5-mini               & 0.118 (98.5) & 0.586 (100.0) & 0.253 (99.0)  & 0.093 (100.0) & 0.569 (100.0) & 0.638 (100.0) & 0.567 (100.0) \\
Gemini-3-flash           & 0.126 (100.0) & 0.560 (100.0) & \textbf{0.463} (100.0) & 0.315 (100.0) & 0.561 (100.0) & 0.693 (100.0) & 0.563 (100.0) \\
TimeOmni-v               & \textbf{0.844} (100.0) & \textbf{0.759} (100.0) & 0.343 (100.0) & \textbf{0.362} (100.0) & \textbf{0.694} (100.0) & 0.606 (100.0) & \textbf{0.598} (100.0) \\
\bottomrule
\end{tabular}}
\end{table}

\begin{table}[h]
\centering
\small
\caption{Forecasting: base MAPE on the parsed subset, with SR (\%) in parentheses. Best per column in \textbf{bold}; on SP500, Gemini-3-flash (4.4503) is marginally below TimeOmni-v (4.4512).}
\label{tab:decomp_fc}
\begin{tabular}{l ccc}
\toprule
Model & PixelRec (200) & SP500 (200) & Terra (200) \\
\midrule
InternVL3.5-4B           & 150.0 (100.0) & 4.76 (100.0) & 296.1 (93.0) \\
InternVL3.5-8B           & 98.0 (100.0)  & 5.44 (100.0) & 156.1 (100.0) \\
Qwen3-VL-8B-Instruct     & 145.5 (100.0) & 5.49 (100.0) & 77.4 (100.0) \\
Qwen3-VL-30B-A3B-Inst.   & 129.2 (100.0) & 4.75 (100.0) & 75.2 (99.5) \\
Qwen2.5-Omni-7B          & 104.4 (100.0) & 7.73 (100.0) & 176.9 (99.5) \\
Qwen3-Omni-30B-A3B-Inst. & 98.0 (100.0)  & 6.72 (100.0) & \textbf{65.4} (100.0) \\
GPT-5-mini               & 91.3 (100.0)  & 4.84 (98.0)  & 100.4 (96.5) \\
Gemini-3-flash           & \textbf{90.8} (99.0) & \textbf{4.45} (100.0) & 95.6 (99.5) \\
TimeOmni-v               & 108.4 (100.0) & 4.45 (100.0) & 159.6 (100.0) \\
\bottomrule
\end{tabular}
\end{table}

\begin{table}[h]
\centering
\caption{Failure taxonomy: \% CUDA OOM / \% other inference failures (e.g., API refusals) / \% unparsable outputs, over all samples. In each cell, SR $=100-$ the sum of the three.}
\label{tab:failure_taxonomy}
\setlength{\tabcolsep}{2.5pt}
\resizebox{\linewidth}{!}{\begin{tabular}{l cccccccccc}
\toprule
Model & AgiBot & CoVLA & CUHK-X & FuFa & HoloA & MIMIC-death & MIMIC-disch & PixelRec & SP500 & Terra \\
\midrule
InternVL3.5-4B           & 0.8/0/71.0   & 0/0/0   & 0/0/0       & 0/0/0   & 0/0/50.7    & 0/0/0 & 0/0/0 & 0/0/0   & 0/0/0   & 0/0/7.0 \\
InternVL3.5-8B           & 1.5/0/79.4   & 0/0/0.2 & 0/0/5.5     & 0/0/2.2 & 0/0/56.3    & 0/0/0 & 0/0/0 & 0/0/0   & 0/0/0   & 0/0/0 \\
Qwen3-VL-8B-Instruct     & 22.1/0/0     & 0/0/0   & 0/0/0       & 0/0/0   & 0/0/5.3     & 0/0/0 & 0/0/0 & 0/0/0   & 0/0/0   & 0/0/0 \\
Qwen3-VL-30B-A3B-Inst.   & 100.0/0/0    & 0/0/0   & 0/0/0.6     & 0/0/0   & 100.0/0/0   & 0/0/0 & 0/0/0 & 0/0/0   & 0/0/0   & 0/0/0.5 \\
Qwen2.5-Omni-7B          & 11.5/0/32.1  & 0/0/0   & 0/0/0       & 0/0/0   & 4.0/0/21.4  & 0/0/0 & 0/0/0 & 0/0/0   & 0/0/0   & 0/0/0.5 \\
Qwen3-Omni-30B-A3B-Inst. & 42.7/0/29.0  & 0/0/0   & 0/0/0       & 0/0/0   & 21.8/0/14.7 & 0/0/0 & 0/0/0 & 0/0/0   & 0/0/0   & 0/0/0 \\
GPT-5-mini               & 0/1.5/0      & 0/0/0   & 0/0.3/0.6   & 0/0/0   & 0/0/0       & 0/0/0 & 0/0/0 & 0/0/0   & 0/0/2.0 & 0/0/3.5 \\
Gemini-3-flash           & 0/0/0        & 0/0/0   & 0/0/0       & 0/0/0   & 0/0/0       & 0/0/0 & 0/0/0 & 0/0/1.0 & 0/0/0   & 0/0/0.5 \\
TimeOmni-v               & 0/0/0        & 0/0/0   & 0/0/0       & 0/0/0   & 0/0/0       & 0/0/0 & 0/0/0 & 0/0/0   & 0/0/0   & 0/0/0 \\
\bottomrule
\end{tabular}}
\end{table}

The taxonomy shows three failure patterns. The open-source Qwen models run out of memory on the long-TS domains (on AgiBot, 100\% for Qwen3-VL-30B-A3B-Instruct, 42.7\% for Qwen3-Omni-30B-A3B-Instruct, 22.1\% for Qwen3-VL-8B-Instruct, and 11.5\% for Qwen2.5-Omni-7B; on HoloAssist, up to 100\%). The InternVL3.5 models stop following the output format on the same domains (71--79\% unparsable on AgiBot, 51--56\% on HoloAssist). The proprietary models rarely fail, but their base accuracy on AgiBot stays low (0.118 and 0.126 Macro-F1). TimeOmni-v has no failure on any test set.

\subsection{Paired Bootstrap Tests}
\label{app:bootstrap}

TimeOmni-v is compared with the strongest baseline of each dataset in Table~\ref{tab:main_results} under a paired sample-level bootstrap with 1{,}000 resamples. $\Delta$ is the swF1 difference for classification (positive favors TimeOmni-v) and the swMAPE difference for forecasting (negative favors TimeOmni-v); $p$ is the one-sided resampling probability that TimeOmni-v is not better, so $p<0.05$ means TimeOmni-v leads significantly and $p>0.95$ means the baseline leads significantly. For MIMIC, $\Delta$ is the mean over the two subtasks, which are resampled separately.

\begin{table}[h]
\centering
\small
\setlength{\tabcolsep}{3pt}
\caption{Paired bootstrap comparison of TimeOmni-v with the strongest baseline per dataset, with two-sided 95\% intervals for $\Delta$.}
\label{tab:bootstrap}
\resizebox{\linewidth}{!}{\begin{tabular}{l ccccccccc}
\toprule
& CoVLA & HoloA & CUHK-X & MIMIC & AgiBot & FuFa & PixelRec & SP500 & Terra \\
\midrule
Baseline & Qwen3-VL-8B & GPT-5-mini & Gemini-3-flash & Qwen3-VL-8B & Gemini-3-flash & Gemini-3-flash & GPT-5-mini & Gemini-3-flash & Qwen3-Omni-30B \\
Test $n$ & 1{,}344 & 945 & 308 & 150 & 131 & 46 & 200 & 200 & 200 \\
$\Delta$ & $+$0.167 & $+$0.125 & $-$0.120 & $-$0.053 & $+$0.717 & $+$0.047 & $+$17.1 & $+$0.00 & $+$93.8 \\
95\% CI  & [$+$0.134, $+$0.199] & [$+$0.084, $+$0.168] & [$-$0.180, $-$0.042] & [$-$0.157, $+$0.049] & [$+$0.630, $+$0.795] & [$-$0.101, $+$0.207] & [$+$3.8, $+$31.0] & [$-$0.76, $+$0.84] & [$+$41.6, $+$169.2] \\
$p$      & $<$0.001 & $<$0.001 & 0.999 & 0.844 & $<$0.001 & 0.246 & 0.995 & 0.522 & $>$0.999 \\
\bottomrule
\end{tabular}}
\end{table}

TimeOmni-v leads significantly on CoVLA, HoloAssist, and AgiBot, the latter already at $n=131$. The strongest baseline leads significantly on CUHK-X, PixelRec, and Terra. On SP500 the narrow interval around zero marks a tie. Future Factories ($n=46$) and MIMIC ($n=150$) remain unresolved, with intervals that span zero, so no ranking conclusion is drawn on them.

\subsection{Inference-Time Modality Removal}
\label{app:inference-ablation}

\begin{figure}[h]
\centering
\includegraphics[width=\linewidth]{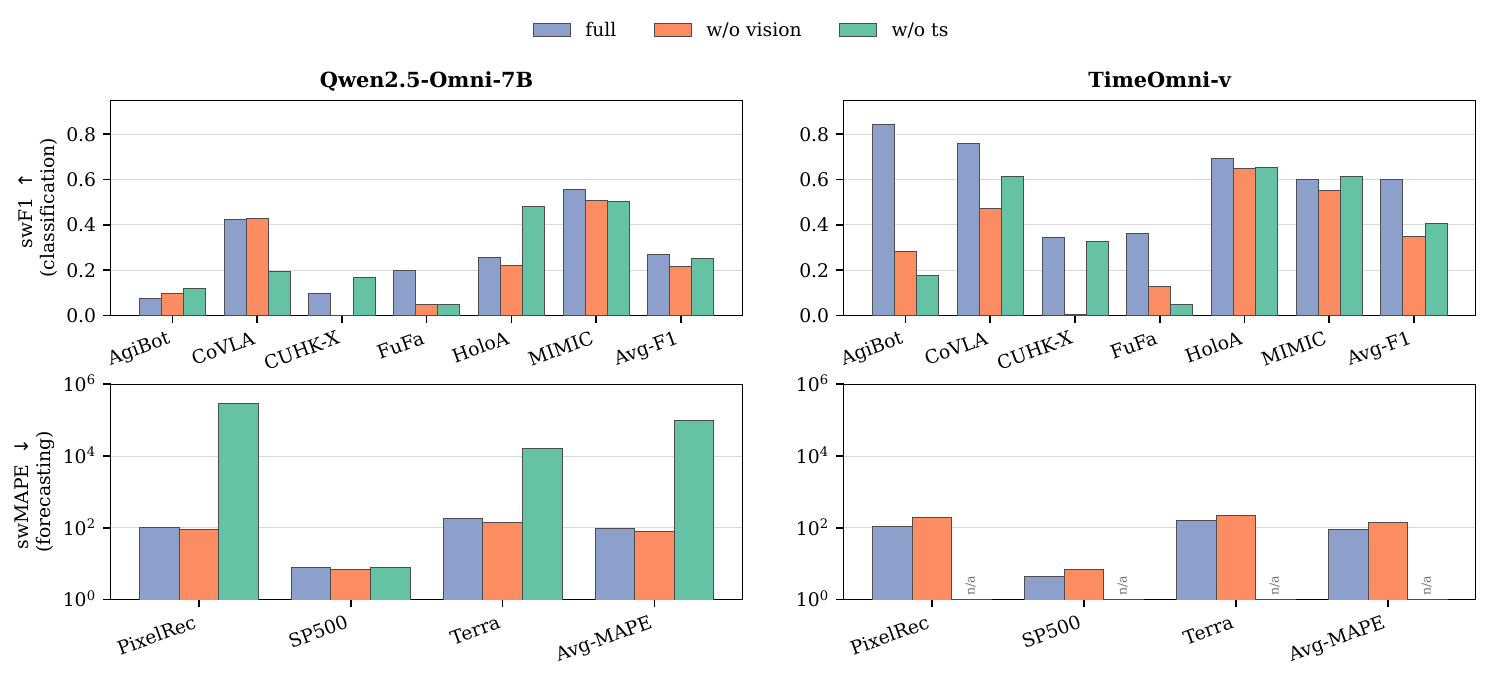}
\caption{Inference-time modality removal under three input conditions: \emph{full} (all modalities provided), \emph{w/o vision} (image/video stream removed), and \emph{w/o ts} (time-series stream removed). Top row: classification swF1 ($\uparrow$); bottom row: forecasting swMAPE ($\downarrow$, log axis); each column is one model. ``n/a'' marks undefined entries (the TimeOmni-v forecast head requires TS input). Avg-F1 and Avg-MAPE aggregate only over defined entries.}
\label{fig:ablation}
\end{figure}

Figure~\ref{fig:ablation} removes one modality at test time from models whose training or evaluation otherwise uses all modalities. Because TimeOmni-v is trained with every modality present, a drop here measures how much the trained model relies on the removed stream, not whether the task needs it; Figure~\ref{fig:modality} answers the second question.

For TimeOmni-v, removing vision lowers Avg-swF1 from 0.601 to 0.349 and removing time series lowers it to 0.406. The three domains outside the training-time ablation show distinct reliance patterns. CUHK-X relies on vision (0.343 to 0.006 without vision, 0.327 without time series), Future Factories relies on time series (0.362 to 0.049 without time series), and HoloAssist changes little under either removal (0.650 without vision, 0.654 without time series). On MIMIC, removing time series at test time raises swF1 from 0.602 to 0.615, in line with the training-time result that the vital signs add little there. Qwen2.5-Omni-7B, which is evaluated zero-shot, often scores higher after a removal (for example, 0.257 to 0.481 on HoloAssist without time series), which indicates that it does not use the removed stream effectively.

\section{Broader Impacts}
\label{app:broader-impacts}

This appendix lists the positive and negative societal impacts that follow from the released artifacts (the MMTA benchmark and the TimeOmni-v codebase). The discussion is scoped to the use cases that MMTA's nine domains directly support; impacts that apply to MLLMs in general but are not specific to this work are not repeated.

\paragraph{Positive impacts.}
MMTA targets domains where decisions are made under joint visual, textual, and temporal evidence -- healthcare, autonomous driving, manufacturing, AR assistance, and climate. A public tri-modal benchmark surfaces failure modes (notably the long-sequence textualized-TS failure mode named in Section~\ref{sec:results-main}) before such systems are placed near deployment, and supports apples-to-apples comparison across model families. Beyond evaluation, TimeOmni-v' compression of the time-series modality lowers the activation-memory and inference-latency barrier to training tri-modal models on commodity GPUs (Section~\ref{sec:results-efficiency}), which makes the modality-aware recipe more accessible to academic and clinical research groups whose hardware budget cannot match those of large industry labs.

\paragraph{Negative impacts and mitigations.}
Three negative impacts are visible. First, the surveillance-style streams in CUHK-X (skeleton-rendered human action), Future Factories (assembly-station cameras), and HoloAssist (egocentric AR video) could in principle be repurposed for unconstrained behaviour monitoring; this risk is bounded by the upstream releases, which already published these streams under their own access policies, and MMTA does not republish raw video beyond what those sources already make available. Second, the MIMIC-derived prognosis tasks (mortality and discharge readiness) inherit demographic biases from the underlying ICU population and are not validated for clinical decision-making; the credentialed-access requirement of PhysioNet remains the primary control, and the benchmark is positioned as a model-evaluation instrument rather than a screening tool. Third, the FinMultiTime-derived forecasting task uses delayed news snippets and historical S\&P 500 closes; the headline numbers are not validated for live trading and should not be read as production-ready signals. Across all three, no fine-tuned model checkpoints from this work are released for sensitive domains (MIMIC, FinMultiTime); only the benchmark splits and the training code are public, which keeps deployment a deliberate downstream choice rather than a default.

\section{Limitations}
\label{app:limitations}

On the benchmark side, MMTA is a collection of tri-modal tasks with diverse cross-modal complementarity patterns, not a benchmark in which every domain strictly satisfies tri-modal non-redundant evidence: CoVLA and CUHK-X carry no sample-level text, FinMultiTime, Terra, and MIMIC combine modalities that share a source (Appendix~\ref{app:data-quality}), and modality necessity is measured by training-time ablation on six of the nine domains (Figure~\ref{fig:modality}). Human validation covers the two generated components, and no per-sample check was run that a sample needs at least two modalities, because a reliable judgment requires domain experts such as clinicians and driving experts. Several test splits are small, and the paired bootstrap leaves the comparisons on Future Factories (46 samples) and MIMIC (150 samples) unresolved.

On the evaluation side, the baselines in Table~\ref{tab:main_results} are evaluated zero-shot, whereas TimeOmni-v is fine-tuned per dataset. The adapted TS-as-text baseline removes this confound only on short-series domains, because TS-as-text fine-tuning runs out of memory on AgiBotWorld and HoloAssist and truncated textual baselines were not trained; on forecasting, the comparison also changes the output from digit decoding to a forecast head. The success-weighted metrics mix semantic errors with parsing and hardware failures, which Appendix~\ref{app:decomposed} separates. The time-series integration has been ported to one other backbone on four datasets, and each domain is trained and evaluated separately, so transfer across backbones and across domains remains untested.

\end{document}